\documentclass[num-refs,twocolumn,serif]{wiley-article}
\usepackage{siunitx}
\usepackage{bm}
\usepackage{stfloats}
\usepackage{graphicx}
\usepackage{caption}
\usepackage{subcaption}
\usepackage{threeparttable}
\usepackage{booktabs}
\usepackage{float}
\usepackage{algorithm}
\usepackage{algorithmic}
\usepackage{multirow}
\usepackage{float}
\usepackage{setspace} 
\usepackage{footnote}
\newcommand{\RNum}[1]{\uppercase\expandafter{\romannumeral #1\relax}}

\newcommand{\Exp}{\mathop{\mathbb E}\displaylimits}
\papertype{Original Article}

\title{Decision-Making under On-Ramp merge Scenarios by Distributional Soft Actor-Critic Algorithm}

\author[1\authfn{1}]{Yiting Kong}
\author[1\authfn{1}]{Yang Guan}
\author[1]{Jingliang Duan}
\author[1]{Shengbo Eben Li}
\author[1]{Qi Sun}
\author[1]{Bingbing Nie}

\contrib[\authfn{1}]{Equally contributing authors.}

\affil[1]{School of Vehicle and Mobility, Tsinghua University, Beijing, 100084, China}

\corraddress{Shengbo Eben Li, School of Vehicle and Mobility, Tsinghua University, Beijing, 100084, China}
\corremail{lishbo@tsinghua.edu.cn}

\fundinginfo{This work is supported by International Science \& Technology Cooperation Program of China under 2019YFE0100200, and also supported by Tsinghua University-Toyota Joint Research Center for AI Technology of Automated Vehicle }

\begin{document}
\begin{frontmatter}
\maketitle
\begin{abstract}
Merging into the highway from the on-ramp is an essential scenario for automated driving. The decision-making under the scenario needs to balance the safety and efficiency performance to optimize a long-term objective, which is challenging due to the dynamic, stochastic, and adversarial characteristics. The Rule-based methods often lead to conservative driving on this task while the learning-based methods have difficulties meeting the safety requirements. In this paper, we propose an RL-based end-to-end decision-making method under a framework of offline training and online correction, called the Shielded Distributional Soft Actor-critic (SDSAC). The SDSAC adopts the policy evaluation with safety consideration and a safety shield parameterized with the barrier function in its offline training and online correction, respectively. These two measures support each other for better safety while not damaging the efficiency performance severely. We verify the SDSAC on an on-ramp merge scenario in simulation. The results show that the SDSAC has the best safety performance compared to baseline algorithms and achieves efficient driving simultaneously.
\end{abstract}
\end{frontmatter}

\section{Introduction}
The demand for improving driving performance has led to the development of intelligent vehicles in recent years, where decision-making is one of the core technologies to realize intelligence \cite{EbenRL}. The ramp is an essential scenario the road but also a complex scene due to its dynamic, stochastic, and adversarial characteristics. Different from the general lane-changing tasks, decision-making under the ramp scenarios must be accomplished in a limited distance with a minimum speed requirement. These characteristics pose great challenges on the development of decision-making algorithm under ramp scenarios, where the safety and efficiency are the two important aspects to be considered. 

Similar to the general lane-changing maneuver, the on-ramp merge maneuver takes immediate actions given the state of surrounding vehicles to achieve an objective related to safety and efficiency. Different from that of the general lane-changing maneuver, the objective optimizes the performance over a long-term horizon. Meanwhile, the goal must be achieved in a limited distance with a speed above the lowest limit, which consequently presents considerable challenges on the balance between the safety and efficiency performance. Several studies has been proposed to solve the decision-making problem under similar scenes, categorized by rule-based and learning-based methods.

Rule-based methods are currently the most common methods and conceptually simple to implement. The gap acceptance theory is the most widely used to model the process of the on-ramp merge problem. In the theory, the vehicle will merge into the mainline only when the distance/time gap is larger than a critical threshold \cite{chen2017decision,rios2016survey}. However, it often leads to conservative behavior in dynamic scenes, because the threshold for behavior selection is difficult to be determined in such scenes. Besides, some studies proposed to select a policy among previously generated candidates by minimizing a cost function of tracking errors and collision avoidance. Such as, Wei \emph{et al.} \cite{wei2013autonomous} assumed the vehicle merges along the center line and get the optimal velocity profile among several candidate strategies. Due to the real time requirement and limited computation power, only a limited number of strategies can be searched and evaluated, leading to a limited performance.

Learning-based methods without hand-crafted rules have also been used to solve similar decision-making problems. A driving model can be learned directly by mimicking drivers’ manipulation using supervised learning (SL) techniques. There are many SL methods, such as decision tree (DT), support vector machine (SVM), and convolution neural network (CNN). In 2001, a study established a  decision-making model to change lane by DT with fuzzy logic \cite{al2001framework}. The parameters of the nodes are trained by minimizing the error with human driver data. The DT can have different decisions in the same situation, and the policy with the largest weight is selected as the final driving policy. In 2017, Vallon et al. trained a lane change model using SVM with features of relative position and relative speed between the ego vehicle and two surrounding vehicles \cite{vallon2017machine}. After the lane-changing behavior is triggered, a lane-changing trajectory with minimum tracking errors is re-planned. The SL methods are also capable of high-dimensional features with the help of deep neural networks. In 2015, NVIDIA's research team established an integrated system using CNN, which successfully realized automated lane change from raw pixels of a single front-facing camera \cite{bojarski2016end}.
Compared to the rule-based methods, these methods are less conservative in dynamic scenes because it is essentially imitations of the human drivers. However, they require massive amounts of natural driving data to cover all possible scenes, which makes the learned policy unsafe to be applied on corner cases \cite{jingliang2019hierarchical}.

Another type of learning-based decision methods is reinforcement learning (RL). Different from SL, RL seeks a driving policy that maximizes long-term returns through trial-and-error, reducing the reliance on driving data \cite{jingliang2019hierarchical, guan2020centralized, guan2019direct, mu2020mixed}. To handle safety issues, one common measure is to consider safety in the policy evaluation, i.e., adding safety terms in the reward. Wang \emph{et al}. established an integrated decision-making model for the first time under the on-ramp merge scenarios using deep Q-networks \cite{wang2017formulation}. They use distance from surrounding vehicles as a reward term to encourage keeping away from them. The method is simple to implement yet has no safety guarantee. Meanwhile, increasing the safety consideration often 
damaging the other performances severely.
Alternatively, some studies consider safety terms as explicit constraints in the policy improvement \cite{duan2019deep, achiam2017constrained}. These methods aim to solve a constrained optimization problem with usually tens of thousand parameters. As a result, not only the algorithms are more complicated, but the solutions are also approximately safe. Another method to deal with safety is the safety shield, which is employed after the training process to further map the policy outputs to the safe action space \cite{mirchevska2018high}. The mapping is done by solving a one-step model predictive control problem with state constraints. It can eliminate unexpected disturbances in the trained policy to prevent from collision, but the optimization problem may gradually become infeasible because of the myopia. The method can fail in that case.

In this paper, we propose an RL-based end-to-end decision-making algorithm, called the Shielded Distributional Soft Actor-Critic (SDSAC), to realize safe and efficient driving at the on-ramp merge scenario. The algorithm balances the performance of safety and efficiency by a framework of offline training and online correction, in which the policy evaluation with safety consideration and the safety shield are both adopted to support each other for better safety performance. In the offline training, the reward is designed with a safety term so that the policy update is guided by a comprehensive evaluation. That reduces the reliance on the safety shield and then the probability of its failure. In the online correction, a safe action is computed from the output of the trained policy by minimizing its distance from the safe action space. To avoid infeasible problems, we control the 
boundary of the safe space using the barrier function technique. The simulation suggests that the SDSAC has the best performance in terms of safety and efficiency compared to baseline algorithms. The contributions are summarised as follows, 1) We propose an easy implemented RL algorithm called SDSAC to boost the safety performance while balancing other performances well. 2) We apply the algorithm on an on-ramp merge scenario with different traffic density in simulation, realizing safe and efficient driving.

The rest of the paper is organized as follows. Section \RNum{2} introduces the preliminaries of RL and our baseline algorithm. Section \RNum{3} introduces the methodology, including the overall framework of our algorithm, offline training and online correction. Section \RNum{4} presents the problem statement and formulation.  Section \RNum{5} presents the experimental settings and implement details, illustrates the results under on-ramp merge scenarios. A brief conclusion of this work is given in the last section \RNum{6}. 

\section{Preliminaries}
In RL, the decision-making problem is described as a Markov decision process (MDP), where the state at the next time step depends only on the state and action at the current time step. The MDP is defined by the tuple $(\mathcal{S},\mathcal{A},R,p)$, where $\mathcal{S}$ denotes the state space, $\mathcal{A}$ denotes the action space, $R(s,a):\mathcal{S}\times\mathcal{A}\rightarrow\mathcal{P}(r)$ is the reward function mapping state-action pairs to a distribution of rewards, $p:\mathcal{S}\times\mathcal{S}\times\mathcal{A}\rightarrow\mathbb{R}$ is the state transition probability. We use $\pi: \mathcal{S}\times\mathcal{A}\rightarrow\mathbb{R}$ to denote a stochastic policy, which maps states to a probability distribution over actions. At each time step $t$, the agent at the state $s_t\in \mathcal{S}$ selects an action $a_t\in \mathcal{A}$. In return, the agent receives the next state $s_{t+1}\in \mathcal{S}$ and a reward $r_t\sim R(s_t,a_t)$. 

In this paper, we employ Distributional Soft Actor-Critic (DSAC) as our baseline algorithm \cite{duan2020addressing}, on which we develop the SDSAC algorithm with high safety performance. DSAC is currently the state-of-the-art model-free RL algorithm. It is based on the maximum entropy learning principle and the value distribution theorem \cite{haarnoja2018softE,Haarnoja2018SACA}. The maximum entropy learning principle improves the exploration ability, while the distributional value improves the
sample complexity and the overestimate error occurred in the value evaluation. In our proposed algorithm, DSAC is applied to update the value and policy network.

\section{Shielded DSAC}
\subsection{Algorithm framework}
\begin{figure*}[htbp]
  \centering%
  \includegraphics[width=0.7\linewidth]{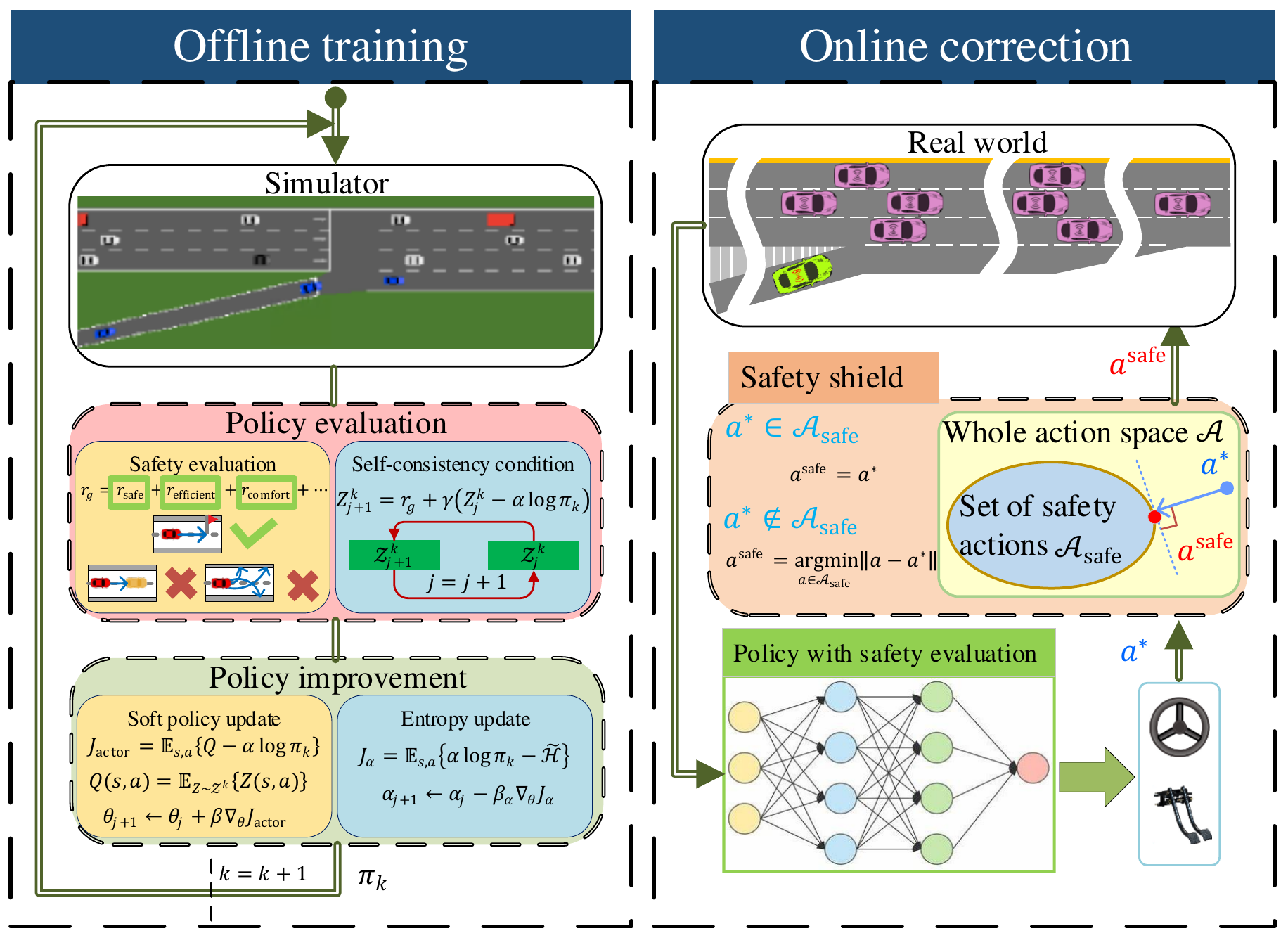}
  \caption{Framework of the SDSAC under on-ramp merge scenarios.}
  \label{fig: framework}
\end{figure*}

SDSAC is designed in a framework of offline training and online correction, as shown in Fig. \ref{fig: framework}. In this framework, both the policy evaluation with safety terms and the safety shield are used to enhance the safety performance while not damaging the efficiency severely. Moreover, automated vehicles need to generate safe decisions in the application to reach goal states without collisions. In contrast, RL methods need to explore in the state space during the training process to learn an accurate value function. This can drive the agent to dangerous states and break the safety requirements. The proposed framework naturally solves the contradiction between safety and exploration by separating these two functionalities.

In the offline training, a policy with comprehensive performance is obtained by alternating steps of policy evaluation and policy improvement. DSAC is employed here for mitigating the value overestimate error to achieve the best asymptotic performance. It is worth noting that the reward function is already designed with a safety term in the policy evaluation, for the propose of boosting safety performance of the trained policy and reducing the failure rate of the safety shield.

The online correction is introduced to further improve the safety. When the well-trained policy is applied online, the safety shield comes after its output to calculate a safe action. Specifically, it maps the action into the nearest action in the safe action space, which is tightened by the barrier function in case no feasible solution can be found. By such a mechanism, the safety performance can be largely enhanced.

\subsection{Offline training with DSAC}\label{section:DSAC}
The long-term return is calculated by the sum of step rewards $r(s,a)$. In DSAC, the reward function is defined as the combination of the general reward $r_{g}$ and the policy entropy term $\mathcal{H}$:
\begin{equation}
r(s,a)= r_{g}(s,a) +\alpha\mathcal{H}(s,a),
\end{equation}
where $\alpha$ adjust the relative importance of the entropy term against the general reward.

In the on-ramp merge task. we design the general step reward $r_{g}$ as:
\begin{equation}\label{eq:rmod}
  r_{g} = \left\{
  \begin{aligned}
    r_{\text{failure}} & \quad \text{failure}  \\
    r_{\text{success}} & \quad \text{success}  \\
    r_{\text{safe}}+r_{\text{efficiency}}+r_{\text{comfort}}+r_{\text{task}} & \quad \text{else}\\
  \end{aligned}\right. ,
\end{equation}
where $r_{\text{failure}}$ is a punishment when the task fails, $r_{\text{success}}$ is a reward when the task is completed successfully, $r_{\text{safe}},r_{\text{efficiency}},r_{\text{comfort}}$ and $r_{\text{task}}$ are respectively the step reward about safety, efficiency, comfort and task completion when the task is underway. Detailed formulation of each term will be introduced in section \ref{section.reward_function}. 

The policy entropy term $\mathcal{H}$ is formulated as:
\begin{equation}
  \mathcal{H}(s,a) = -\alpha \log \pi(a|s),
\end{equation}
which is inversely related to the action selection probability. The action with lower probability has higher uncertainty, so it is assigned with higher reward to be selected with higher probability in the future. 

The objective of DSAC is to learn a policy that maximizes the expected long-term return:
\begin{equation}
\label{eq.policy_objective}
J_{\pi} = \mathop{\mathbb{E}}_{(s_i,a_i)\sim\rho_{\pi},r\sim R}\Big[\sum^{\infty}_{i=t}\gamma^{i-t} [r_{g}(s_i,a_i)+\alpha\mathcal{H}(\pi(a_i|s_i))]\Big],
\end{equation}
where $\rho_{\pi}$ is the state-action distribution induced by policy $\pi$ in environment, $\gamma \in (0,1]$ is a discount factor. 

The stochastic policy of DSAC is evaluated by a special state-action value function:
\begin{equation}
Z(s_t,a_t) = r_g(s_t,a_t) +\sum^{\infty}_{i=t+1}\gamma^{i-t} \{r_g+\mathcal{H}\},
\end{equation}
where $Z(s_t,a_t)$
is the random variable of the long-term return from a state-action pair $(s_t,a_t)$. The corresponding variant of Bellman operator is derived as:
\begin{equation}\label{eq.soft_distri_bellman}
\mathcal{T}^{\pi}_{\mathcal{D}}Z(s,a) \overset{D}{=}r_g+\gamma( Z^{\pi}(s',a')-\alpha \log\pi(a'|s')),
\end{equation}
where $A\overset{D}{=}B$ denotes that two random variables A and B have equal probability laws and  $(s',a')$ denotes the random state-action pair in the next time step.

Instead of learning the expected value of $Z(s,a)$, i.e., Q-values, DSAC directly learns its distribution to evaluate the stochastic policy $\pi$. The distribution of $Z(s,a)$ is denoted as $\mathcal{Z}^{\pi}(Z(s,a)|s,a): \mathcal{S}\times\mathcal{A}\rightarrow \mathcal{P}(Z(s,a))$, which is a mapping from $(s,a)$ to distributions
over soft state-action returns.

The function approximation is necessary for solving large-scale continuous problems. In DSAC, the stochastic policy $\pi_{\theta}(a|s)$ and the value distribution $\mathcal{Z}_{\omega}(\cdot|s,a)$ are parameterized as Gaussian distribution, where the mean and variance are given by neural networks with parameters $\theta$, $\omega$. In order to stabilize the learning process, the corresponding target networks with separate parameters $\theta'$, $\omega'$ are introduced. 

In the policy evaluation step, we minimize the KL divergence between the target return distribution and the current return distribution. The objective is formulated as:
\begin{equation}
J_{\mathcal{Z}}(w) =  \mathop{\mathbb{E}}_{(s,a)\sim \rho_{\pi}}\big[D_{\rm{KL}}(\mathcal{T}^{\pi}_{\mathcal{D}}\mathcal{Z}_{\text{old}}(\cdot|s,a),\mathcal{Z}_{w}(\cdot|s,a))\big]
\end{equation}
Its parameter is updated using the gradient decent: 
\begin{equation}
 w \leftarrow w-\beta_{\mathcal{Z}}\nabla_w J_{\mathcal{Z}}(w),
\end{equation}
where $\beta_{\mathcal{Z}}$ is the learning rate of the value distribution network.

In the policy improvement step, the policy is updated by maximizing the parameterized objective \eqref{eq.policy_objective}, which can be rewritten as
\begin{equation}
    J_{\text{actor}}(\theta) =\Exp_{\substack{(s,a)\sim \rho_\pi}}\Big[Q(s,a)-\alpha\log(\pi_{\theta}(a|s))\Big],
\end{equation}
where $Q(s,a)=\Exp_{Z(s,a)\sim\mathcal{Z}_{w}(\cdot|s,a)}[Z(s,a)]$ and the parameter $\theta$ is updated by: 
\begin{equation}
\theta \leftarrow \theta + \beta  \nabla_\theta J_{\text{actor}}(\theta),
\end{equation}
where $\beta$ is the learning rate of the policy network.

The target networks use a slow-moving update rate,
parameterized by $\tau$:
\begin{equation}
  \begin{aligned}
    w' &\leftarrow  \epsilon w+(1-\tau)w'\\
    \theta' &\leftarrow  \epsilon \theta+(1-\tau)\theta'
  \end{aligned}
\end{equation}

The temperature $\alpha$  is updated by minimizing the
following objective:
\begin{equation}
\begin{aligned}
J(\alpha) &=\mathop{\mathbb{E}}_{(s,a)\sim\rho_{\pi}}[-\alpha \log\pi_{\theta}(a|s)-\alpha\overline{\mathcal{H}}]\\
\alpha &\leftarrow \alpha- \beta_{\alpha}\nabla_{\alpha} J(\alpha),
\end{aligned}
\end{equation}
where $\overline{\mathcal{H}}$ is the expected entropy. The detailed derivations of the gradients can be found in \cite{duan2020addressing}. 

\begin{algorithm}[!htb]
\caption{SDSAC Algorithm}
\label{alg:DSAC}
\begin{algorithmic}
\STATE \textbf{1.Offline Traing}
\STATE Initialize parameters $\omega$, $\theta$, and $\alpha$
\STATE Initialize target parameters $\omega'\leftarrow\omega$, $\theta'\leftarrow\theta$
\STATE Initialize learning rate $\beta_{\mathcal{Z}}$, $\beta$, $\beta_{\alpha}$ and $\tau$ 

\REPEAT
\STATE Select action $a\sim\pi_{\theta}(a|s)$
\STATE Calculate reward $r$ with safety evaluation
\STATE Observe new state $s'$
\STATE Store transition tuple $(s,a,r,s')$ in buffer $\mathcal{B}$
\STATE
\STATE Sample $N$ transitions $(s,a,r,s')$ from $\mathcal{B}$
\STATE Update return distribution $\omega \leftarrow \omega - \beta_{\mathcal{Z}}\overline{\nabla_{\omega}J_{\mathcal{Z}}(\omega)}$
\STATE Update policy $\theta \leftarrow \theta + \beta\nabla_{\theta} J_{\theta}(\theta)$
\STATE Adjust temperature $\alpha \leftarrow \alpha - \beta_{\alpha}\nabla_{\alpha} J(\alpha)$
\STATE Update target networks:
\STATE \qquad $\omega' \leftarrow  \tau \omega+(1-\tau)\omega'$
\STATE \qquad $\theta' \leftarrow  \tau\theta+(1-\tau)\theta'$
\UNTIL Convergence  
\STATE \textbf{2.Online Application}
\STATE Select action using the trained policy  $a^*_t = \mathbb{E}_a\{\pi_{\theta}(a|s_t)\}$ 
\IF{$h(f(s_t, a^*_t))\leq (1-\lambda)h(s_{t})$}
\STATE Take action $a^{\text{safe}}_t=a^*_t$ 
\ELSE
\STATE $\arg\min ||a-a_t^*||^2$
\STATE $s.t. h(f(s_t,a))\leq (1-\lambda)h(s_{t})$
\STATE Take action $a^{\text{safe}}_t=a$ 
\ENDIF
\end{algorithmic}
\end{algorithm}
\subsection{Online action correction with state constraints}
\begin{figure}[hpbt]
  \centering%
  \includegraphics[width=0.7\linewidth]{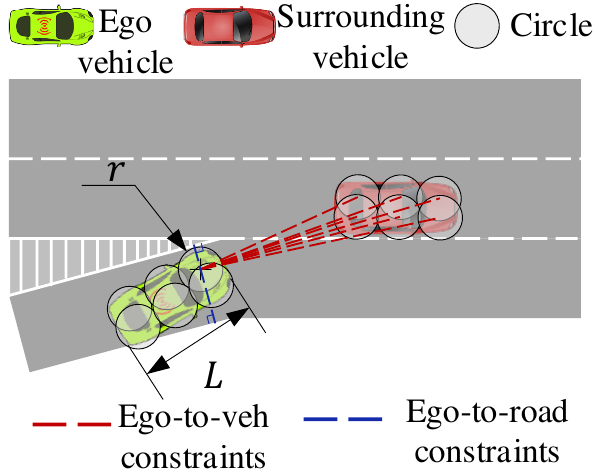}
  \caption{Representation of state constraints.}
  \label{fig: state_cons}
\end{figure}

Applying the trained policy directly may cause dangerous moves due to the lack of rigid safety guarantee. The online action correction is necessary to further improve the safety performance. The online action correction finds the nearest actions in the safe space $\mathcal{A}_{\text{safe}}$, which is formulated as a QP problem:
\begin{equation}\label{eq:QP}
  a^{\text{safe}}_t = \left\{
  \begin{aligned}
    &a^*_t, \quad \text{if}\quad a^*_t \in \mathcal{A}_{\text{safe}} \\
    &\arg\min\limits_{a} ||a-a_t^*||^2 , \quad \text{else}\\
    &s.t. a \in\mathcal{A}_{\text{safe}}
  \end{aligned}\right.
\end{equation}
where $a^*_t$ is the policy output.
The safe action in $\mathcal{A}_{\text{safe}}$ should guarantee the next model predictive state $\hat{s}_{t+1}$ is safe, i.e., collision-free with the surrounding vehicles and road edges, where $\hat{s}_{t+1} = f(s_t,a_t)$ and $f(\cdot)$ is the vehicle dynamics and the motion prediction model. Therefore, the QP problem constrains $\hat{s}_{t+1}$ by keeping the distance from obstacles  larger than a safe distance. 
To determine the safe distance, all vehicles are represented by six circles as illustrated in Fig. \ref{fig: state_cons}. 
The circle radius is denoted as $r =\frac{L}{6}$, where $L$ is the vehicle length. As a result, each circle of the ego vehicle has six constraints with a surrounding vehicle and two constraints with the left/right road edges. Considering $N$ of surrounding vehicles, we have $6\cdot(6N+2)$ constraints at each step. The state constraints between the ego and surrounding vehicles are denoted as $h(\hat{s}_{t+1})_{ij}\leq 0$. The state constraints between the ego and the road edges are denoted as $h(\hat{s}_{t+1})_{ik}\leq 0$, where $i\in [1,...,6]$ is the index of the ego circles, $j\in[1,...,6N]$ is the index of the surrounding vehicle circles, $k\in[1,2]$ is the index of road edges. There are 36$N$ ego-to-vehicle constraints in total. 
\begin{equation}
h(\hat{s}_{t+1})_{ij} = r_i + r_j-\sqrt{(x_i-x_j)^2+(y_i-y_j)^2}\leq 0,
\end{equation}
where $x_i,y_i,r_j$ are the positions and the radius of the center of circle $i$, $x_j,y_j,r_j$ are the circle center and the radius of the surrounding vehicle $j$. There are 12 ego-to-road constraints.
\begin{equation}
h(\hat{s}_{t+1})_{ik}= r_i -\sqrt{(x_i-x_k)^2+(y_i-y_k)^2}\leq 0,
\end{equation}
where $x_k,y_k$ are the nearest points on both sides of the road edge. In the following, we use $h(\cdot)$ to denote the collection of $h(\cdot)_{ij}$ and $h(\cdot)_{ik}$.

\begin{figure}[thpb]
  \centering%
  \includegraphics[width=0.7\linewidth]{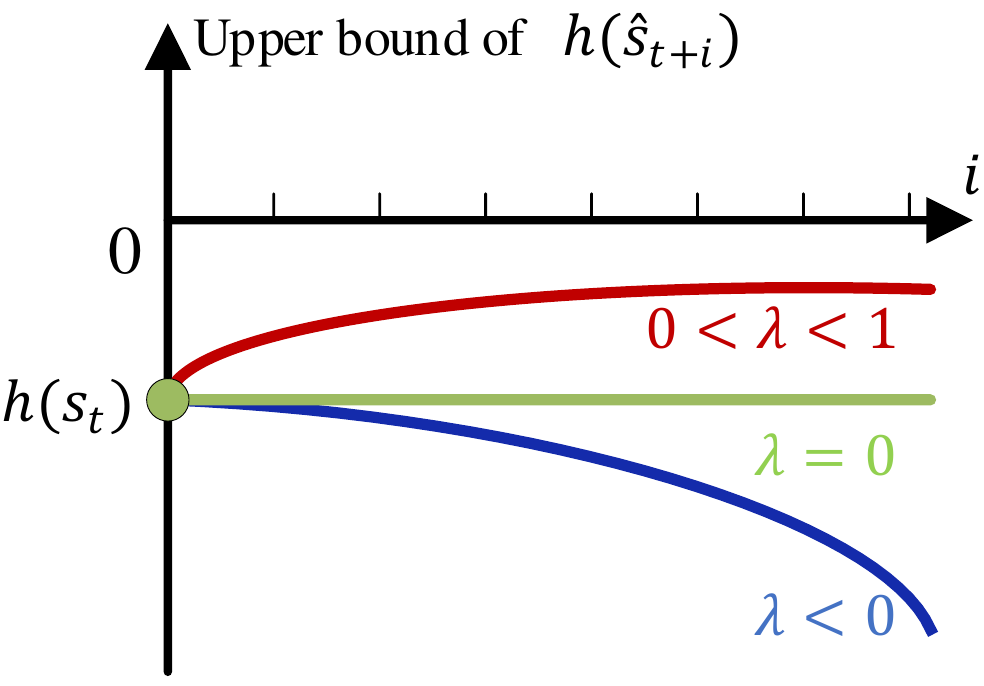}
  \caption{The state constraints under barrier function.}
  \label{fig: bf}
\end{figure}

This online action correction only considers the state constraints in one step. If the QP problem directly uses the state constrains above, the optimal safe action could be aggressive. Even worse, the problem may become infeasible because of the myopia. In this paper, we use the barrier function technique to constrain the variation trend of the state constraints so that the corrected action can always be found. The transformed state constraint becomes
\begin{equation}\label{eq:bf}
\Delta h(\hat{s}_{t+1}) + \lambda h(s_t) \leq 0  , \forall t \in \{0,...,\infty\},
\end{equation}
where $\Delta h(\hat{s}_{t+1}) =  h(\hat{s}_{t+1})- h(s_t)$, $\lambda$ adjusts the level of conservation. Equation (\ref{eq:bf}) reveals that 
\begin{equation}
\begin{aligned}
h(\hat{s}_{t+i}) \leq (1-\lambda)^ih(s_t)\leq0, i \in \{1,...\infty\} 
\end{aligned}
\end{equation}
That is, the barrier function results in more rigorous constraints. As illustrated in Fig. \ref{fig: bf}, the upper bound becomes $(1-\lambda)^ih(s_t)$, which can be controlled by $\lambda$. And the smaller $\lambda$ is, the more rigorous the constraints are. Such a mechanism prevents the ego vehicle from getting into states that are infeasible, reducing the failure rate of the safety shield.

The procedure of SDSAC is shown in Algorithm \ref{alg:DSAC}.

\section{Problem statement and formulation}
\begin{figure}[thpb]
  \centering%
  \includegraphics[width=1\linewidth]{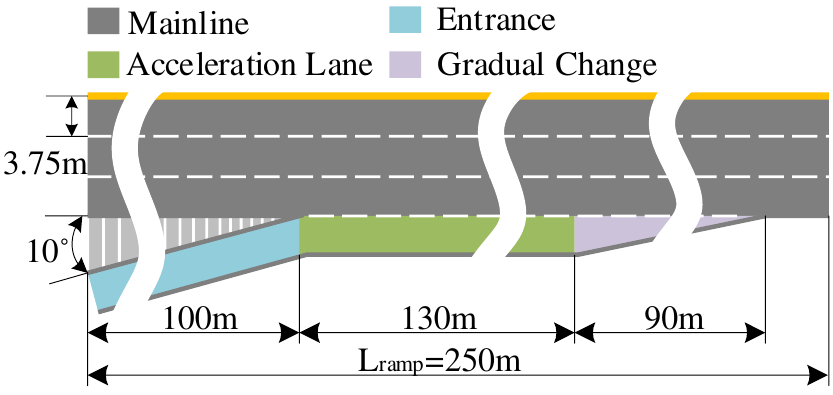} 
  \caption{On-ramp merge scenario design.}
  \label{fig: task_ego}
\end{figure}

In this paper, we focus on a typical on-ramp merge scenario shown in Fig. \ref{fig: task_ego}. The mainline is a three-lane highway with lane width of $3.75$ m. The number and types of vehicles in the mainline are stochastic. The ego vehicle is initialized on the ramp with random states. A success merge happens when the ego vehicle enters the mainline with a small heading angle before it reaches the end of the acceleration lane. So in this problem setting, the destination is not specific.

This problem can be formulated as an RL problem by defining the state space, action space and reward function. In the training process, the state information and reward collected from simulations are used to update the value and policy networks by back-propagation. Then, the updated policy is used to explore in the environment to collect data for the next iteration. In the online correction phase, the well trained policy outputs actions depending on the states, while the network parameters do not change any more. And the output action needs to go through the shield to be mapped to a safe action before used in the real world.

\subsection{State and action space}
The state space is defined as a vector:
\begin{equation}
s = [s_{\text{ego}},s_{\text{veh}}],
\end{equation}
which includes both the information of ego vehicle and surrounding vehicles. 

\begin{figure}[thpb]
  \centering%
  \includegraphics[width=0.7\linewidth]{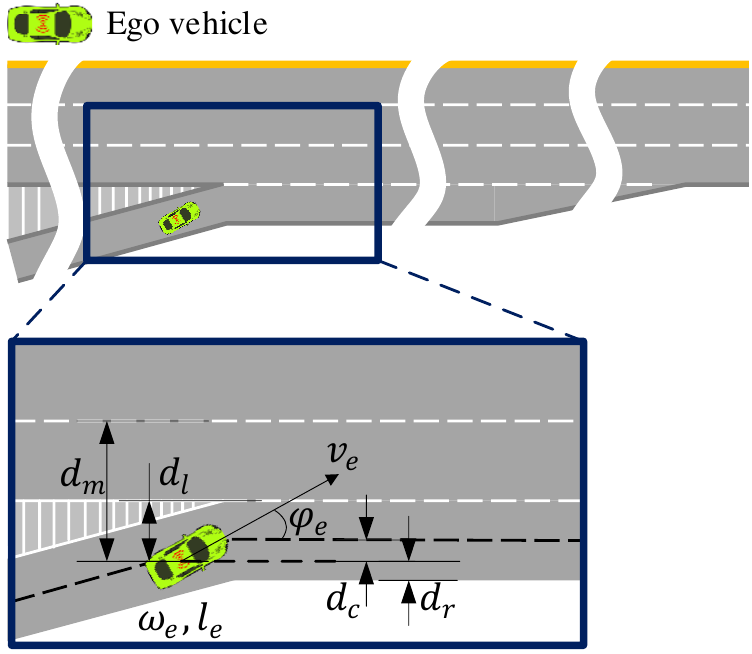} 
  \caption{The state information about the ego vehicle.}
  \label{fig: s_ego}
\end{figure}

The information about the ego vehicle $s_{\text{ego}}$ is formulated as:
\begin{equation}
s_{\text{ego}} =[v_e,w_e,l_e,\phi_e, d_c,d_l,d_r,d_m]
\end{equation}
As illustrated in Fig.\ref{fig: s_ego}, $v_e$ denotes the velocity, $w_e,l_e$ are the width and the length, $\phi_e$ is the heading angle, $d_c,d_l,d_r$ are the distance to the road center-line, the left and right road boundary respectively, and $d_m$ is the lateral distance to the mainline.

\begin{figure}[thpb]
  \centering%
  \includegraphics[width=0.7\linewidth]{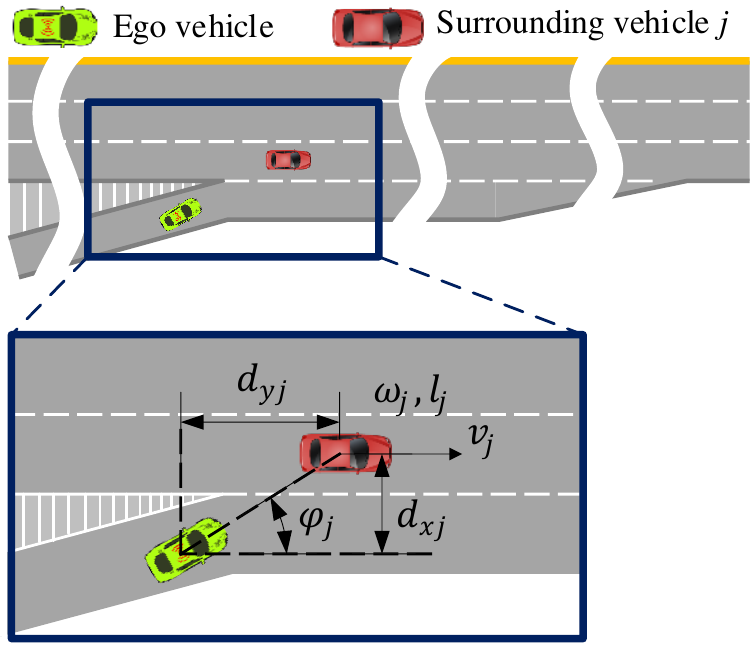} 
  \caption{The state information about the surrounding vehicles.}
  \label{fig: s_veh}
\end{figure}

The information about surrounding vehicles $s_{\text{veh}}$ is denoted as:
\begin{equation}
s_{\text{veh}} = [\text{veh}_1,...,\text{veh}_j,...,\text{veh}_8] ,
\end{equation}
which consists of the nearest leading and following vehicles in the lane where the ego vehicle is on and the two adjacent lanes. The information of a surrounding vehicle is formatted as:
\begin{equation}
\text{veh}_j = [v_j,\varphi_j,w_j,l_j,d_{xj},d_{yj}]
\end{equation}
where $v_j$ denotes the velocity, $\varphi_j$ denotes its heading angle, $w_j,l_j$ are the width and length, $d_{xj},d_{yj}$ are the lateral and longitudinal distance to the ego vehicle as shown in Fig. \ref{fig: s_veh}. 

For action $a$, we choose acceleration $a_x$ and the front wheel angle $\delta$ of the ego vehicle. Considering the vehicle physical constraints, we restrain the acceleration and the front wheel angle in certain ranges. In
total, 56-dimensional continuous state space and 2-dimensional continuous action space are
constructed.

\subsection{Reward function}\label{section.reward_function}
As introduced in section \ref{section:DSAC}, the reward function is a combination of general reward $r_g$ and policy entropy term. As shown in (2), the general reward $r_g$ is composed of three kinds of rewards. When a success merge happens, the agent receives a large positive reward 1000 to encourage this situation. On the contrary, when the agent fails to complete the task, i.e., a collision or timeout happens, a large negative reward -200 is given as punishment. Otherwise, the task continues and the
step reward is designed under consideration of safety, task
completion, efficiency, and comfort. 

To be specific, $r_{\text{safe}}$ is formulated as:
\begin{equation}
\begin{aligned}
  r_{\text{safe}} &= k_{s}(d_{\text{safe}}-d_v)\\
  d_v &= \min(\sqrt{d_{xj}^2+d_{yj}^2)}, j \in{1,\cdots,8},
\end{aligned}
\end{equation}
where $k_{s}$ is a negative parameter used to adjust the safety importance, $d_{\text{safe}}$ is the safe distance computed by the sum of the radius of two circles, and $d_v$ is the minimum distance from the surrounding vehicles. If $d_v>d_{\text{safe}}$, it will receive a positive reward or otherwise a penalty.

The reward $r_{\text{task}}$ reflects the performance of task completion. In this problem, a success merge is not the only requirement. The lane-keeping performance before the merge is also considered, which is formulated as:
\begin{equation}\label{eq.reward_t}
  r_{\text{task}} = k_{t1}d_m^2+k_{t2}\varphi_e^2+k_{t3}d_c^2,
\end{equation}
where $k_{t1},k_{t2},k_{t3}$ are negative weights. The reward guides the ego vehicle merging into the target mainline and keeping consistent with the center line of its current road in terms of its position and heading angle.  

To enhance efficiency, the ego is encouraged to drive as the expected speed. Then, the reward on efficiency term is defined as: 
\begin{equation}
  r_{\text{efficiency}} = k_e(v_e-v_{\text{exp}}),
\end{equation}
where $v_{\text{exp}}$ is the expected speed, The weight $k_e$ is a negative value.

The reward about comfort term is defined as: 
\begin{equation}
  r_{\text{comfort}} = k_{c1}\Delta{\varphi}_e^2+k_{c2}\delta^2+k_{c3}a_x^2,
\end{equation}
where $\Delta{\varphi}_e$ is the yaw rate, and the comfort weights $k_{c1},k_{cs},k_{c3}$ are negative.

The weights in each term are shown in Table. \ref{table.weights}.
\begin{table}[htbp]
\caption{Parameter design in the reward function.}
\label{table.weights}
\centering
\begin{tabular}{p{0.018\textwidth}p{0.018\textwidth}p{0.028\textwidth}p{0.028\textwidth}p{0.018\textwidth}p{0.018\textwidth}p{0.028\textwidth}p{0.028\textwidth}p{0.028\textwidth}}
\hline
$k_{s}$ & $k_{t1}$ & $k_{t2}$ & $k_{t3}$& $k_{e}$ & $v_{\text{exp}}$ & $k_{c1}$ & $k_{c2}$& $k_{c3}$\\
-3 & -1 & -20 & -20& 0.5&15& -15 &-15& -15\\

\hline
\end{tabular}
\end{table}

\section{Experiments}
\subsection{Experimental settings}
The proposed decision-making method is trained and tested in the environment that we set up based on a commonly used traffic simulator, SUMO \cite{lopez2018microscopic}. The on-ramp and the mainline are build by a graphical network editor incorporated in SUMO. Each lane emits a determined number of vehicles every second. The vehicle starts at a determined start position with a random start speed and stays near a random desired speed, which is given by a normal distribution among a fleet of the vehicle. Note that the vehicle speed is still capped at the speed limit for different vehicle types. The surrounding vehicles are controlled by the incorporated car-following model and lane-changing model to avoid collisions with each other. 

The position of the ego vehicle is updated by a dynamic model with provided action commands. Even though the surrounding vehicles are aware of the ego vehicle, the collision is still not avoidable if the ego vehicle takes unreasonable actions. 

\subsection{Implementation details}
In our problem setting, the mainline is 320 meters long with a speed limit of 35 m/s and intersected by the on-ramp at 100m with an angle of $10^{\circ}$. There are four types surrounding vehicles in the traffic and each type has a different driving behavior, cooperative or adversarial. The system frequency is set to be 10 Hz. The acceleration and the front wheel angle can be any real value within the range $a_x\in[-3,3]\rm{m/s^2}$, $\delta\in[-0.7,0.7]\rm{rad}$.

The SDSAC is trained in a Parallel Asynchronous Buffer-Actor-Learner architecture, where 6 learners, 6 actors and 4 buffers are designed to accelerate the learning speed. Both the value function and  policy use multiple layers perception with 5 hidden layers as approximate functions, consisting of 256 units per layer, with Gaussian Error Linear Units (GELU) between each layer. The Adam method with a cosine annealing learning rate is used as optimizer to update all the parameters. The hyperparameters are listed in Table \ref{table.hyper}.

\begin{table}[htbp]
\caption{Detailed hyperparameters.}
\label{table.hyper}
\centering%
\vskip 0.15in
\begin{tabular}{c|c}
\hline
\textbf{Hyperparameters} & \textbf{Value} \\
\hline
Hidden units  & 256 \\\hline
Hidden layers & 5\\\hline
Hidden layers activation&  GELU\\\hline
Optimizer type &  Adam\\\hline
Adam parameter & $\beta_{1}=0.9, \beta_{2}=0.999$\\\hline
Actor learning rate &$5{\rm{e-}}5\rightarrow5{\rm{e-}}6 $\\\hline
Critic learning rate & $1{\rm{e-}}4\rightarrow1{\rm{e-}}5 $\\\hline
$\alpha$ learning rate &$5{\rm{e-}}5\rightarrow5{\rm{e-}}6 $\\\hline
Discount factor $\gamma$ & 0.99\\\hline
Target update rate $\tau$ & 0.001\\\hline
Update delay $m$& 2\\\hline
Reward scale & 5\\\hline
Actor number& 6\\\hline
Learner number & 6\\\hline
Buffer number & 4 \\ \hline
Max steps per episode & 1000\\
\hline
\end{tabular}
\vskip -0.1in
\end{table}

\subsection{Results and discussion}
Three sets of experiments are implemented in this section to verify the effectiveness of the proposed SDSAC algorithm. The first one is to verify the function of the safety evaluation by comparing the training results of the DSAC with/without the safety evaluation when the online correction is disabled. The second is to study the impact of the online action correction on the performance by the comparison of SDSAC and DSAC. And the last is functionality verification of the SDSAC algorithm, in which two simulations conducted by it are visualized and analyzed.

\subsubsection{Comparison of the DSAC with/without the safety evaluation}
\begin{figure}[h]
  \centering%
  \includegraphics[width=1\linewidth]{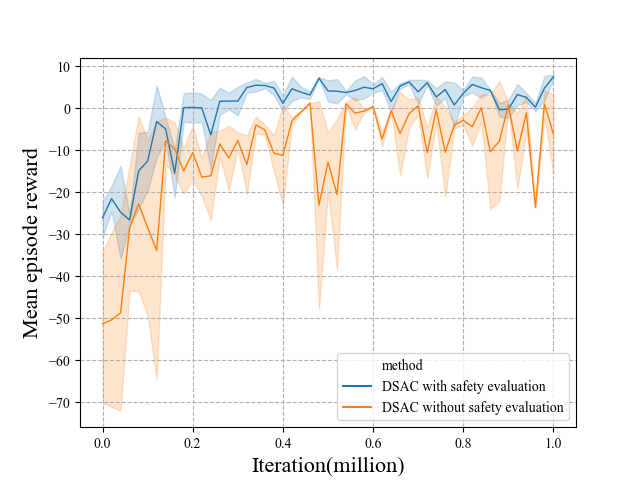} 
  \caption{Training curves of the DSAC with/without safety evaluation.}
  \label{fig: mean episode reward}
\end{figure}

In this section, we disable the online action correction of the SDSAC, and explore the impact of the safety evaluation on the offline training and online application. Specifically, we train two policy networks by DSAC with/without $r_{\text{safe}}$ term in the reward, and show their performances during the training process. Each of them is trained over 5 runs with different random seeds, and each run takes one million iterations. The average return over the best 3 of 5 episodes without exploration noise is used to evaluate the policy every 20000 iterations. As shown in Fig. \ref{fig: mean episode reward}, the solid line demonstrates the mean episode reward, and the shaded area is the confidence interval of $95\%$ over 5 runs. With an obvious uptrend of rewards, the ego vehicle learns good policies nearly after 0.4 million iterations. The results suggest that the policy trained by the SDSAC with safety evaluation has much smaller variance compared to that of the SDSAC without safety evaluation, which means the safety evaluation leads to a more stable and safer policy.

To further verify its impact on online applications, we disable the online action correction and run 10000 trials under on-ramp merge scenarios for each of the trained policy to compare them statistically. A successful merge is counted when the ego vehicle gets into the highway without collision within 30 seconds. Otherwise, it would be counted as a fail one because of the collision or timeout. As shown in Table. \ref{table.comparasion}, the safety consideration $r_{\text{safe}}$ in policy evaluation can largely improve the success rate of the trained policy by reducing collisions.

\subsubsection{Comparison of the SDSAC and DSAC}
In this experiment, we aim to verify the effect of the online action correction by comparing the statistic performance of the SDSAC and DSAC.
The DSAC directly takes the trained policy's output, while the SDSAC corrects the action using the safety shield with a barrier function paramaterized by $\lambda$. We vary the parameter $\lambda \in \{0.1, 0.3, 0.5, 0.9\}$ and run 10000 trails for each $\lambda$. We report the success rate and the failure rate in Table. \ref{table.comparasion}. The results show that the online action correction improves the safety performance significantly (86.65\% $\rightarrow$ 94.42\%), where the majority of the improvement benefits from the sharp drop of the collision rate (11.03\% $\rightarrow$ 0.87\%), and small sacrifice of the efficiency (2.32\% $\rightarrow$ 4.71\%). Besides, when we decrease the $\lambda$, the safety performance increase accordingly while the driving efficiency decrease monotonically. Especially when $\lambda=0.1$, the collision number per million kilometers can be reduced to three, which is 1/10 of that in $\lambda=0.9$, but the efficiency is also halved. This is consistent with the theoretical results that the smaller the $\lambda$ takes, the more conservative of the ego vehicle. Therefore, the $\lambda$ can be served as a micro regulator in SDSAC for the balance of the safety and efficiency performance.

Overall, the comparison results demonstrate that both the safety evaluation in offline training and the action correction with state constraints in online applications matter in SDSAC. By taking these two measures, SDSAC can boost the safety performance while not damaging the efficiency severely compared to the baseline algorithm, realizing safe and efficient driving in the on-ramp merge scenario.

\begin{table*}[t]\label{tb:comparasion}
\centering
\captionsetup{justification=centering}
\caption{Statistical comparisons for three different methods in online application.}
\label{table.comparasion}
\begin{threeparttable}
\begin{tabular}{p{0.30\textwidth}p{0.13\textwidth}|p{0.13\textwidth}|p{0.18\textwidth}|p{0.1\textwidth}}
\hline
    \multirow{2}*{Method} & & \multirow{2}*{Success} & \multicolumn{2}{c}{Failure}\\
    \cline{4-5}           & & & Collision\tnote{$\dagger$} & Timeout \\ 
\hline
\multirow{4}{0.3\textwidth}{SDSAC} &$\lambda$= 0.1& $92.82\%$ & $0.11\%\ (3.1)$  & $7.07\%$ \\
  &$\lambda$= 0.3& $93.79\%$ & $0.28\%\  (9.3)$  & $5.93\%$ \\
 &$\lambda$= 0.5& $94.58\%$ & $0.47\%\  (15.7)$ & $4.95\%$ \\
 &$\lambda$= 0.9& $94.42\%$ & $0.87\%\  (29.2)$ & $4.71\%$ \\
\hline
  \multicolumn{2}{l|}{DSAC with safety evaluation} & $86.65\%$ &$11.03\%\  (318.4)$ & $2.32\%$ \\
\hline
 \multicolumn{2}{l|}{DSAC without safety evaluation} & $75.19\%$ &$22.38\%\  (629.8)$ & $2.43\%$ \\
\hline
\end{tabular}
\begin{tablenotes}
\item[$\dagger$] The value in the bracket is the equivalent collision number per million kilometers. 
\end{tablenotes}
\end{threeparttable}
\end{table*}

\subsubsection{Demonstrations of the policy under SDSAC}
To verify the functionality and adaptability of SDSAC in dynamic scenes, we demonstrate the driving performance in two simulation cases with different traffic densities. For each case, we display the results by snapshots during the simulation. Besides, the key states and actions are visualized, including the velocity, heading angle, acceleration and front wheel angle.

\begin{figure}[!htbp]
  \includegraphics[width=1\linewidth]{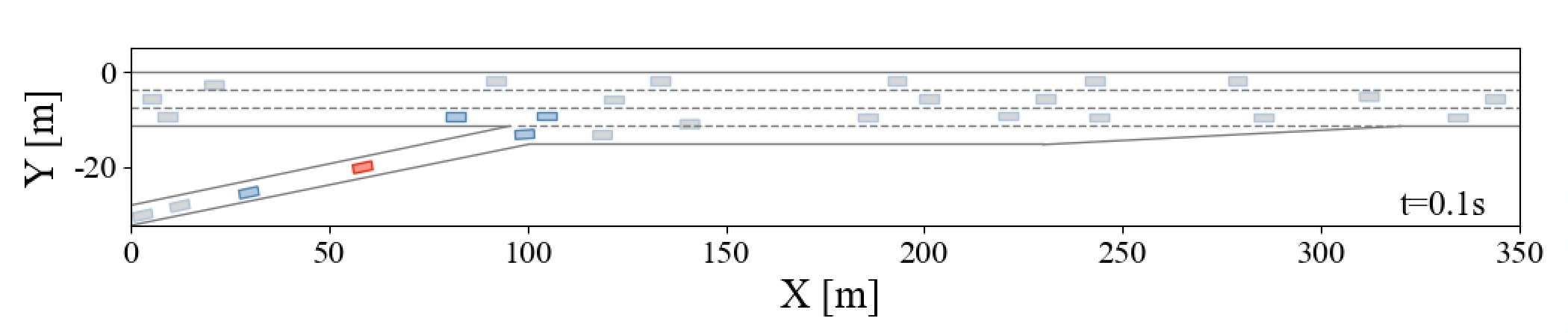}  
  \includegraphics[width=1\linewidth]{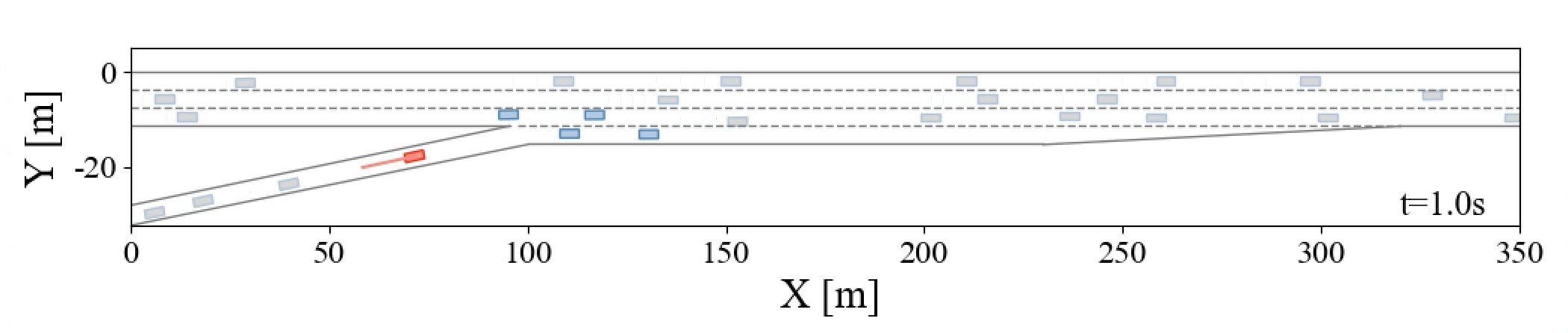}  
  \includegraphics[width=1\linewidth]{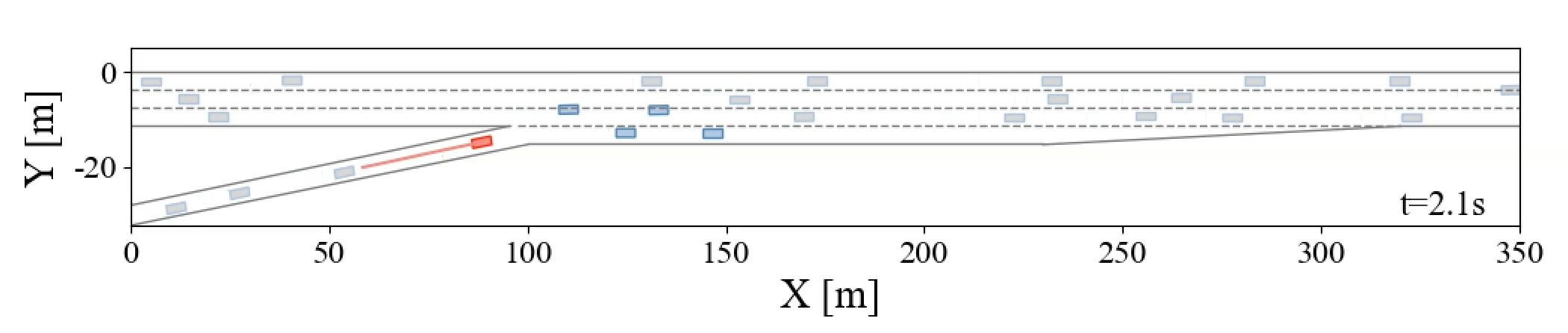}  
  \includegraphics[width=1\linewidth]{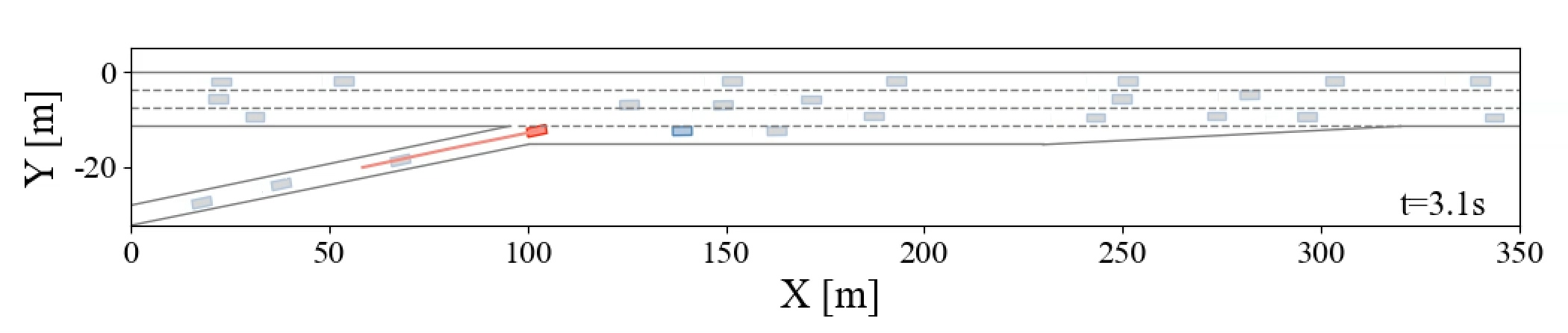}  
  \includegraphics[width=1\linewidth]{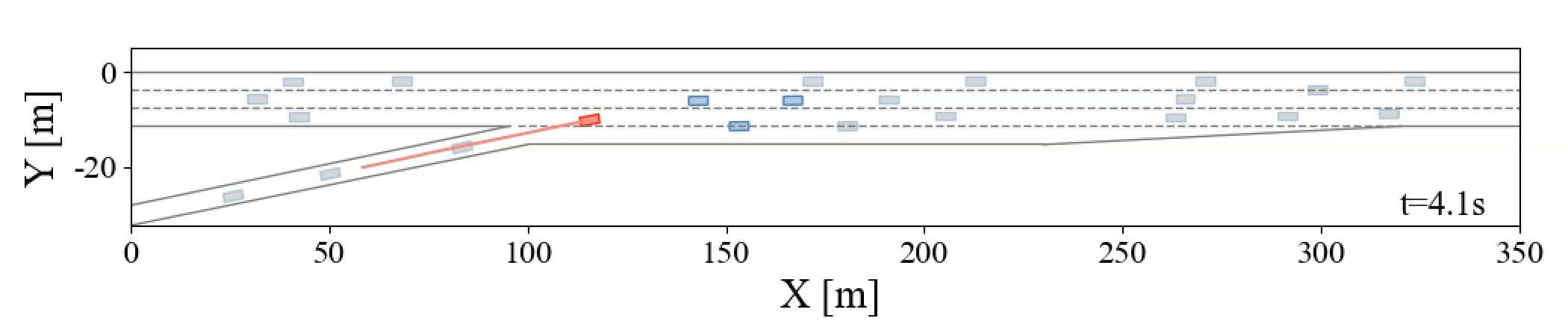}  
  \includegraphics[width=1\linewidth]{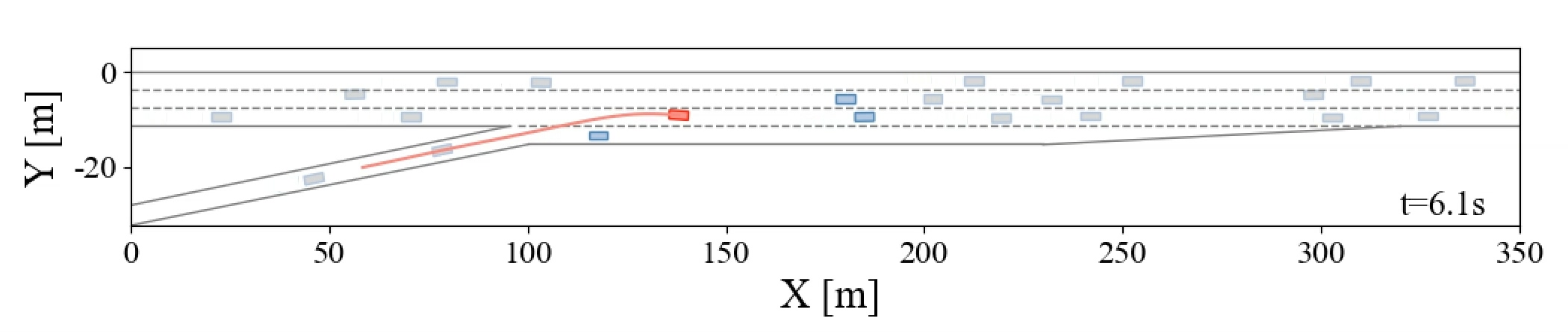}  
  \includegraphics[width=1\linewidth]{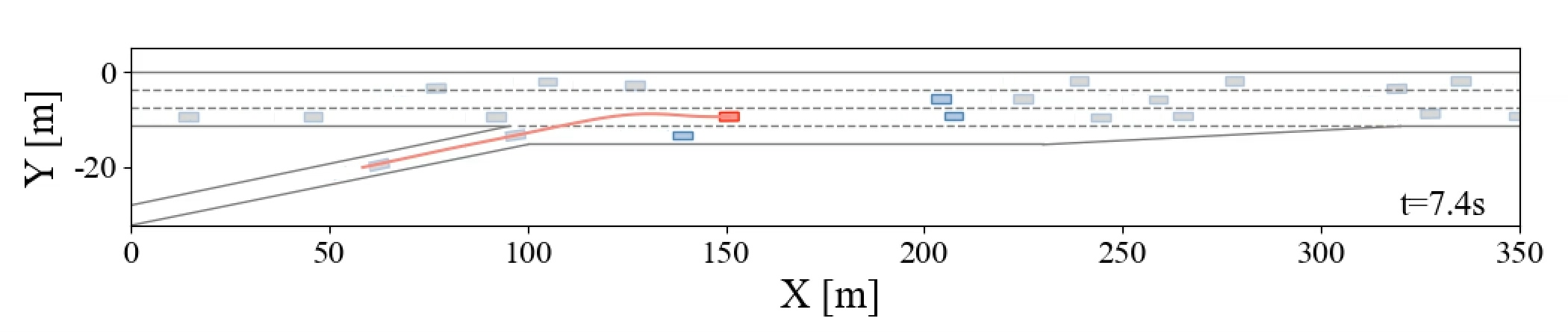}
  \caption{Demonstration of simulation case 1.}
  \label{fig:Sim1_Tra}
\end{figure}

\begin{figure}[!htbp]
\centering
\captionsetup[subfigure]{justification=centering}
\subfloat[Velocity]{\label{fig.velocity1}\includegraphics[width=0.22\textwidth]{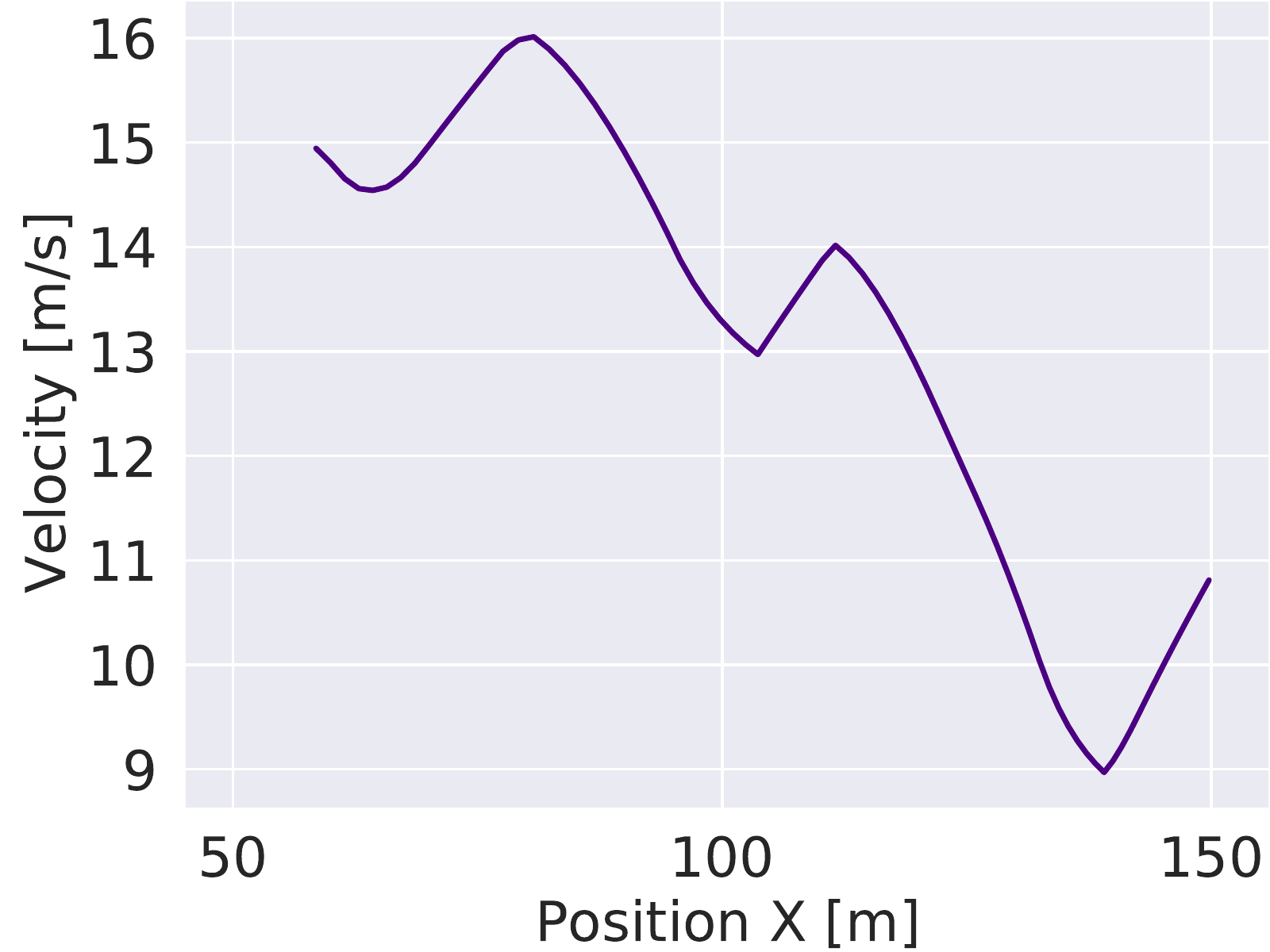}}
\subfloat[Heading angle]{\label{fig.heading1}\includegraphics[width=0.22\textwidth]{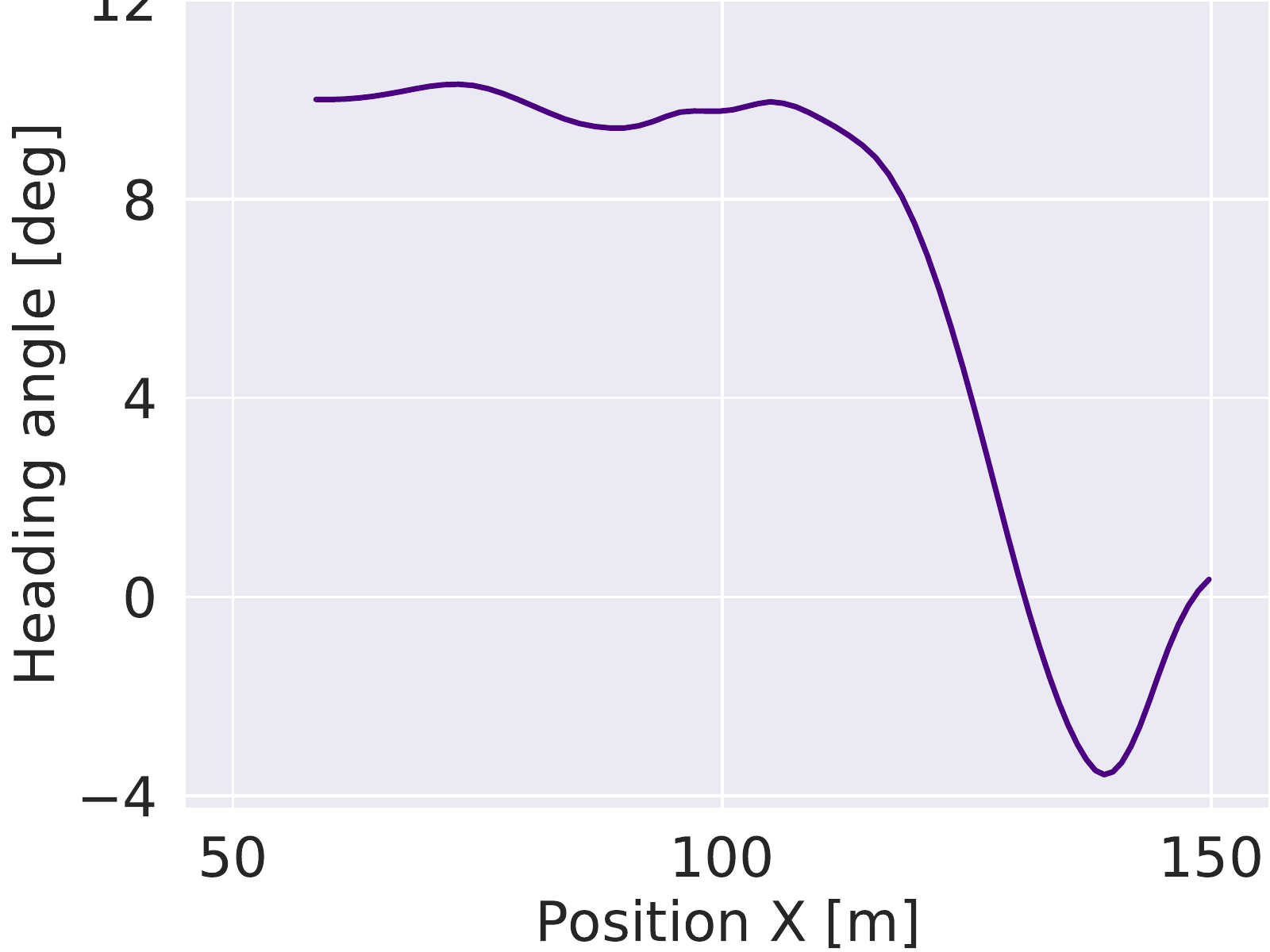}}\\
\subfloat[Acceleration]{\label{fig.acc1}\includegraphics[width=0.22\textwidth]{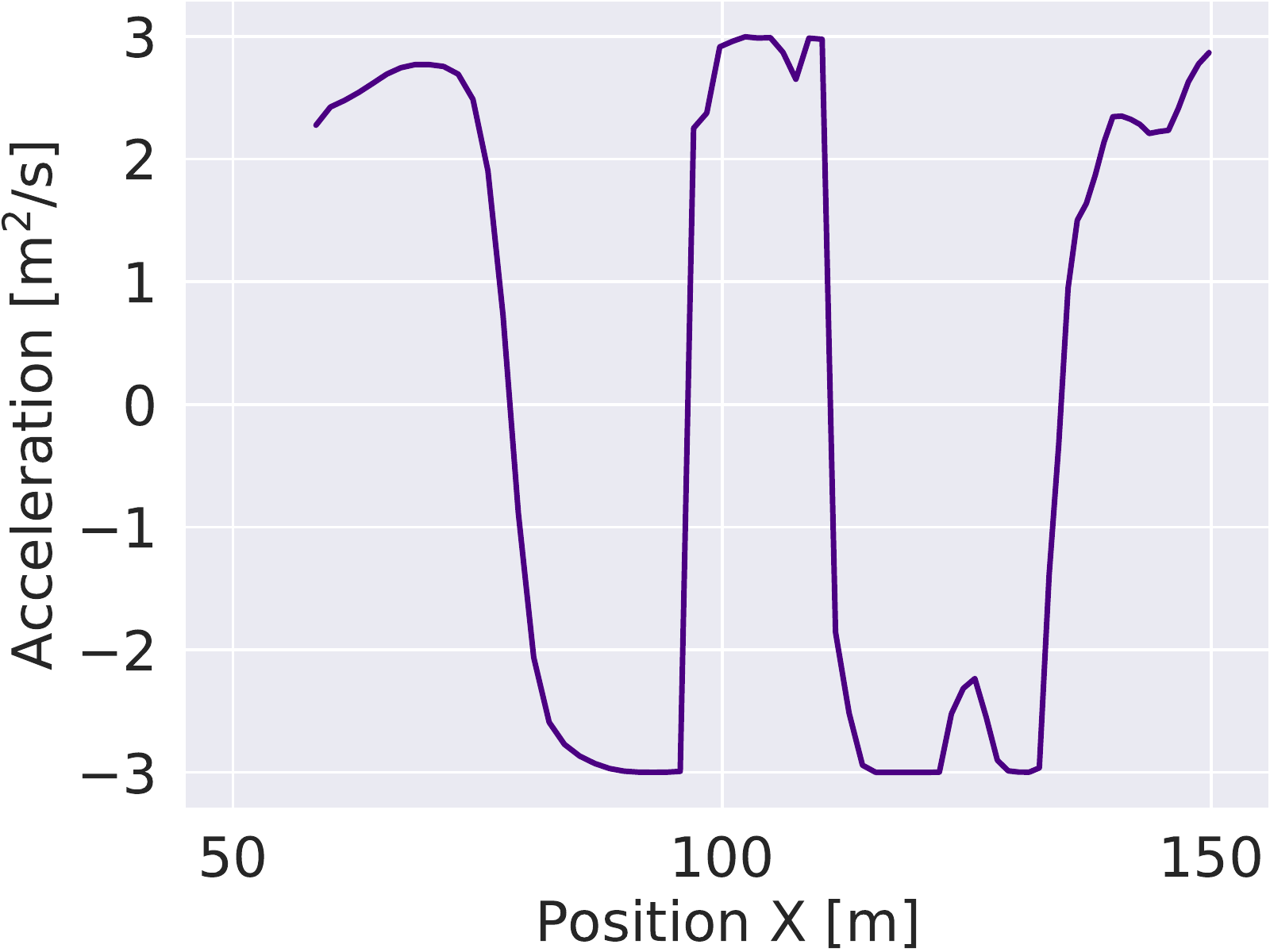}}
\subfloat[Front wheel angle]{\label{fig.delta1}\includegraphics[width=0.22\textwidth]{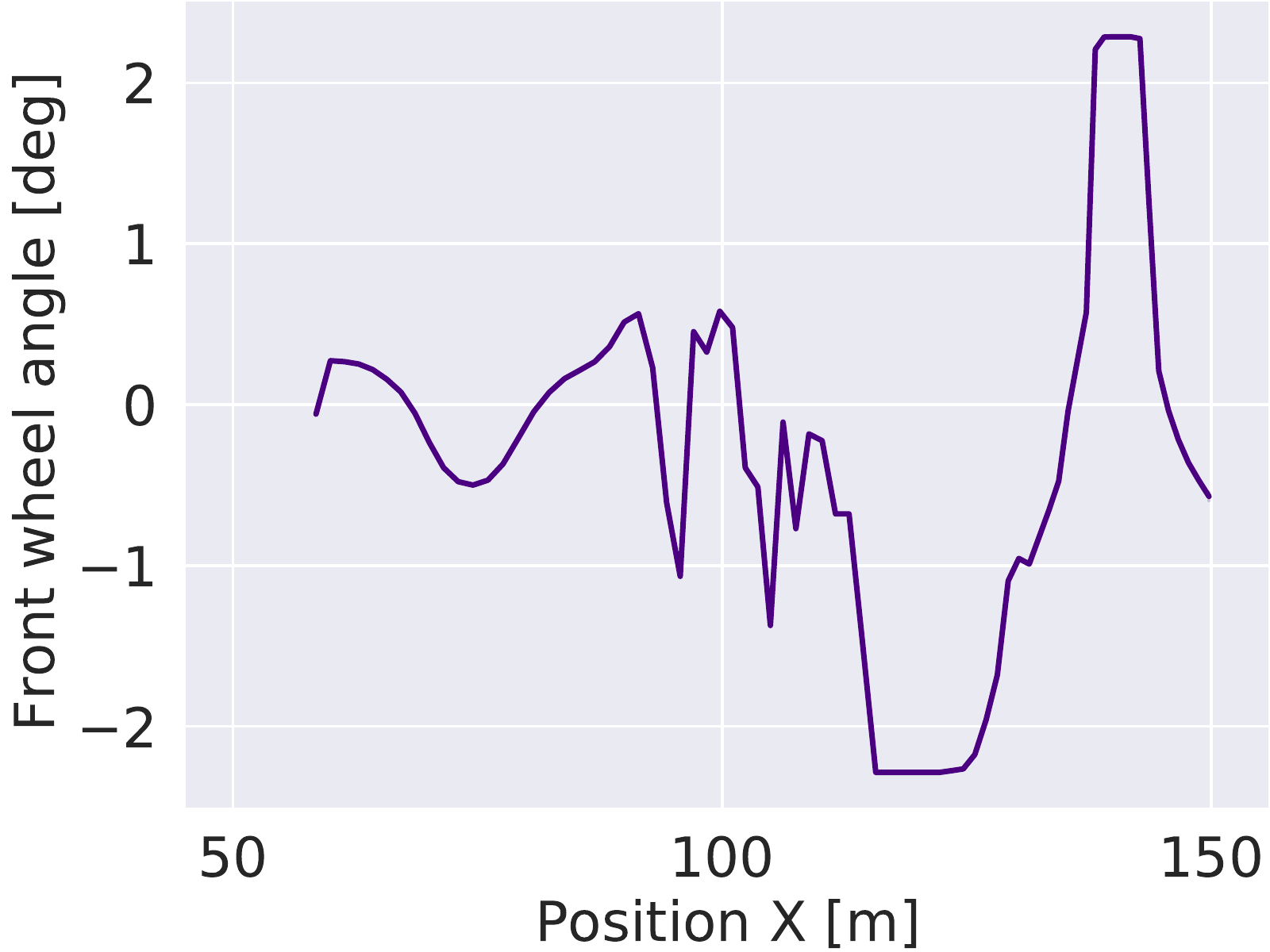}}
\caption{States and actions in simulation case 1.}
  \label{fig:Sim1_state_action}
\end{figure}

In the first simulation, the traffic density is set to be sparse, as shown in Fig. \ref{fig:Sim1_Tra} and Fig. \ref{fig:Sim1_state_action}. The ego vehicle slows down on the ramp to keep away from the front vehicles for the consideration of safety. Before it goes out of the ramp, the front vehicles in the mainline are not interested anymore. Therefore, the ego starts to accelerate to the expected speed to maximize the efficiency.  Once entering the acceleration lane, there is no surrounding vehicles in the mainline, so the ego vehicle turns left to the mainline immediately to complete the task and avoid the timeout failure. After merging into the mainline, the front vehicles are considered again so that the ego decelerate to keep a distance from the them.

\begin{figure}[!htbp]
  \includegraphics[width=1\linewidth]{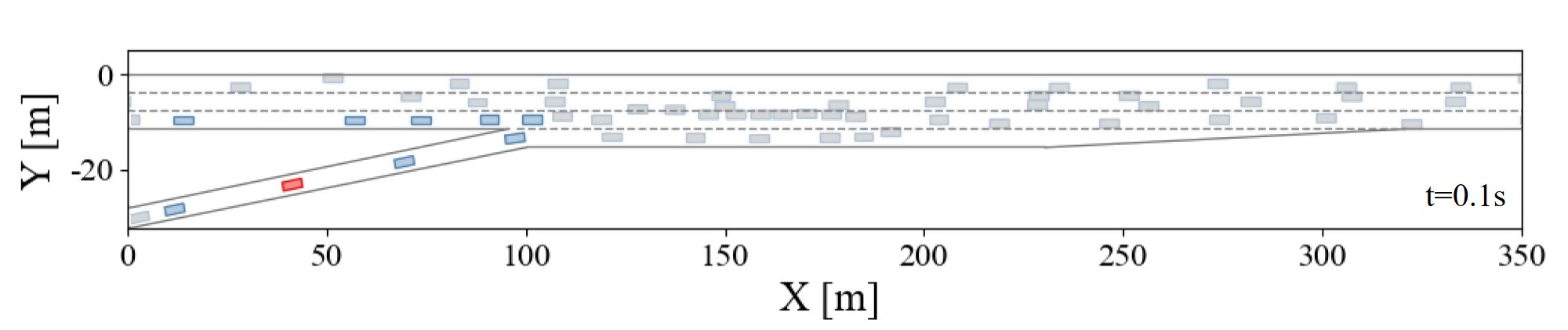}  
  \includegraphics[width=1\linewidth]{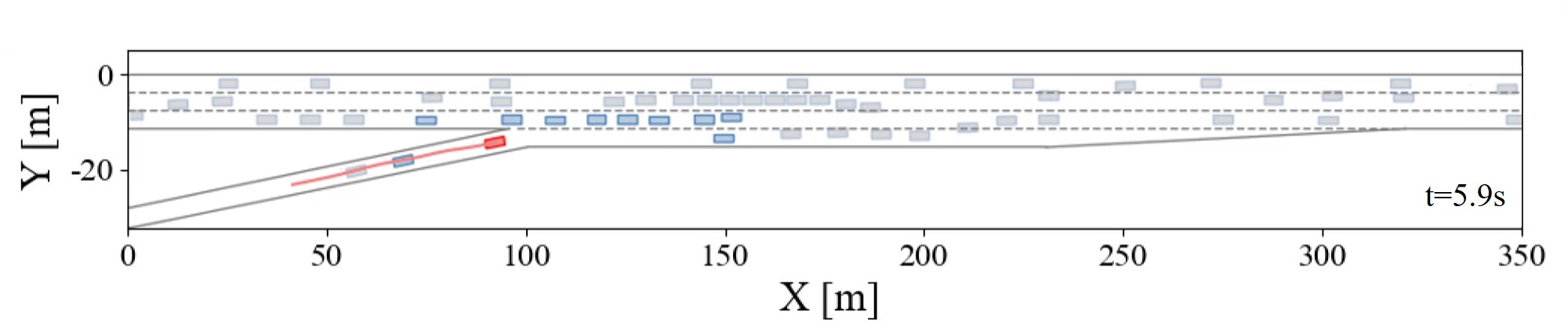}  
  \includegraphics[width=1\linewidth]{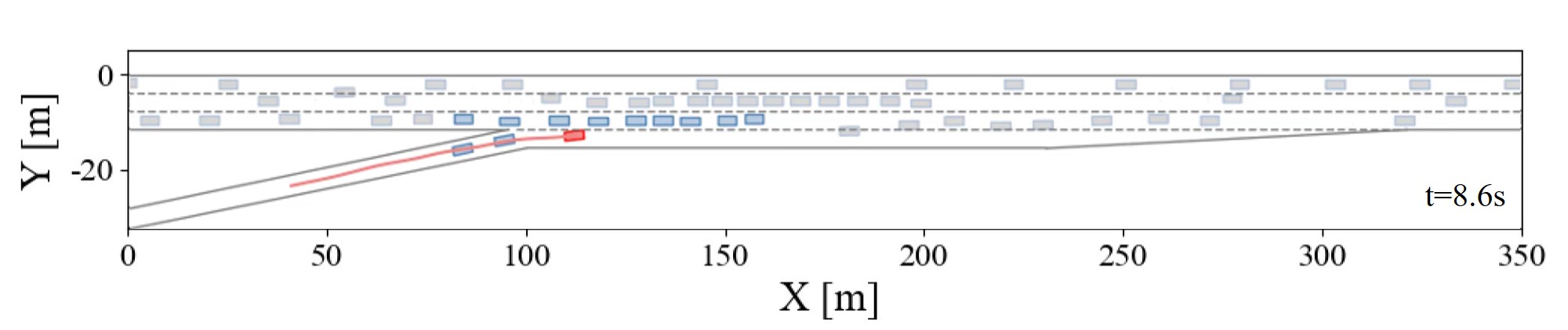} 
  \includegraphics[width=1\linewidth]{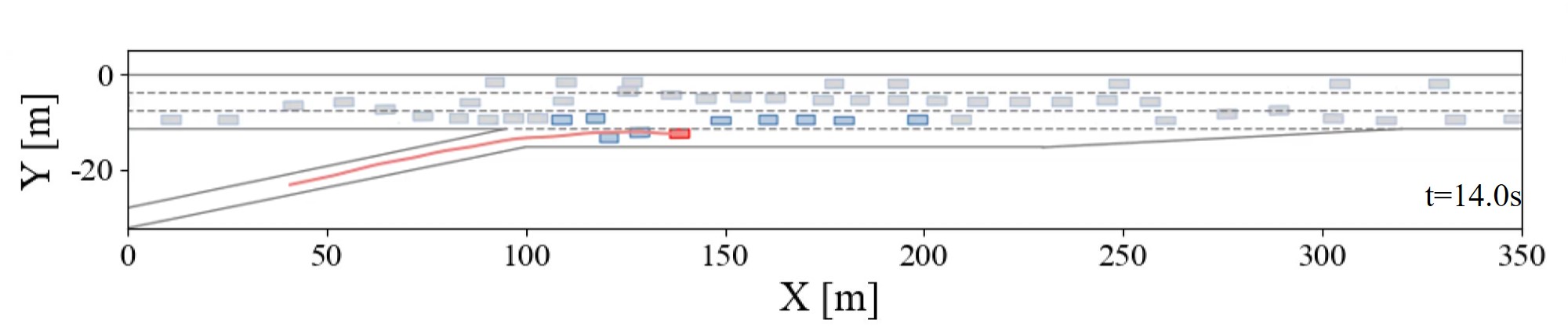} 
  \includegraphics[width=1\linewidth]{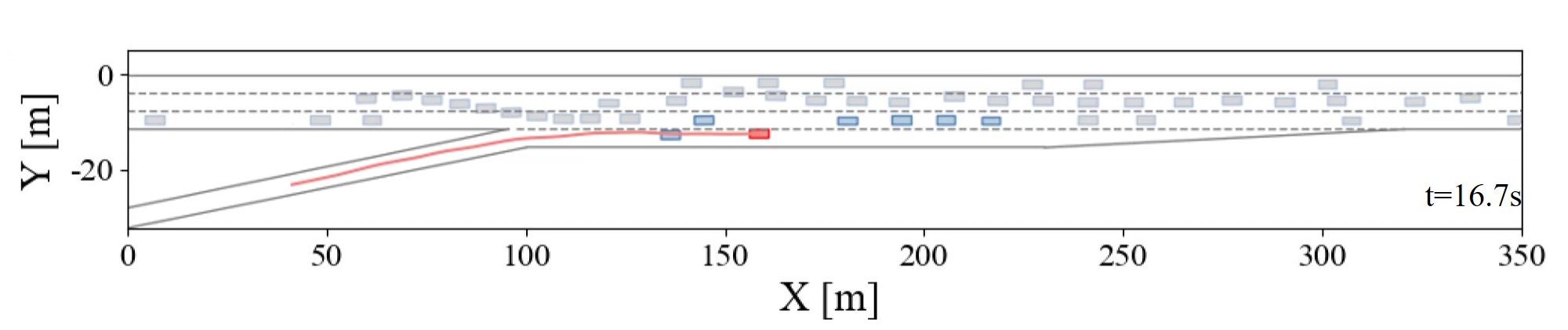} 
  \includegraphics[width=1\linewidth]{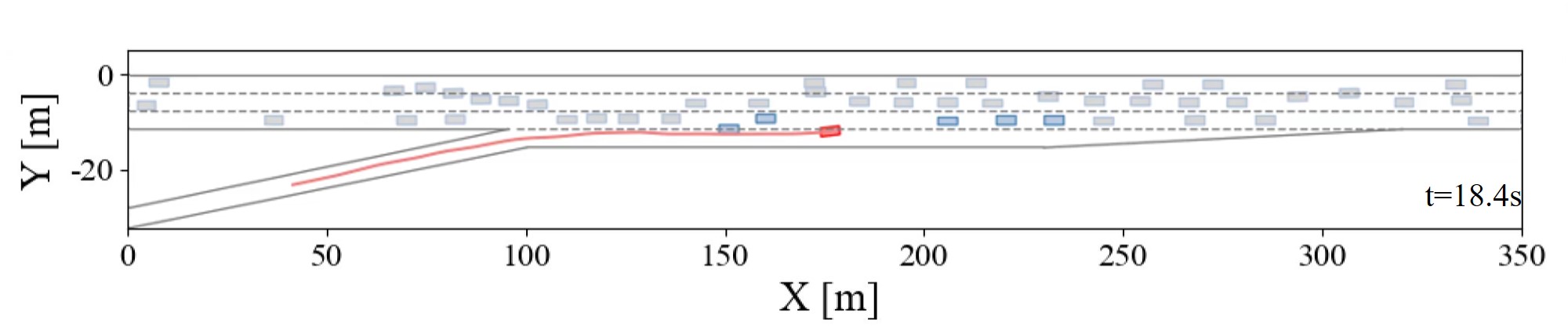} 
  \includegraphics[width=1\linewidth]{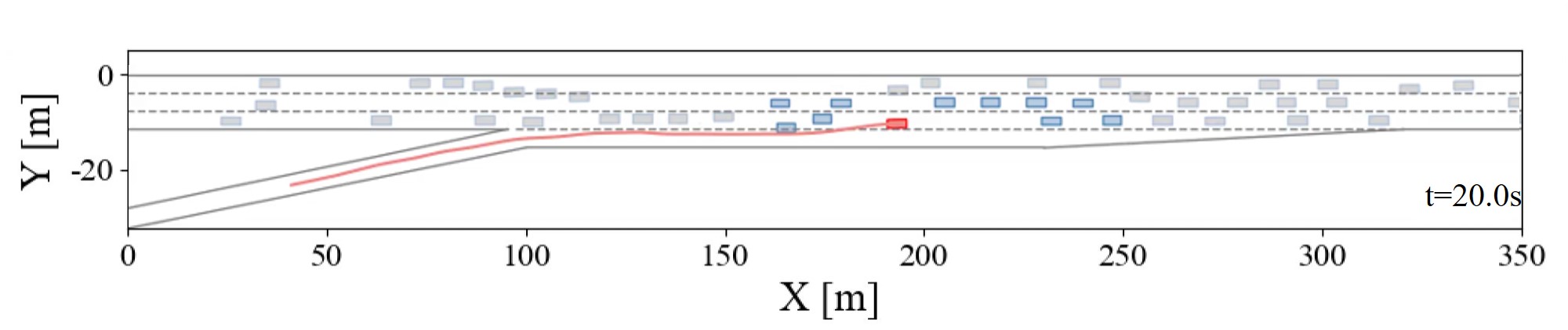}  
  \caption{Demonstration of simulation case 2.}
  \label{fig:Sim2_Tra}
\end{figure}

\begin{figure}[!htbp]
\centering
\captionsetup[subfigure]{justification=centering}
\subfloat[Velocity]{\label{fig.velocity2}\includegraphics[width=0.22\textwidth]{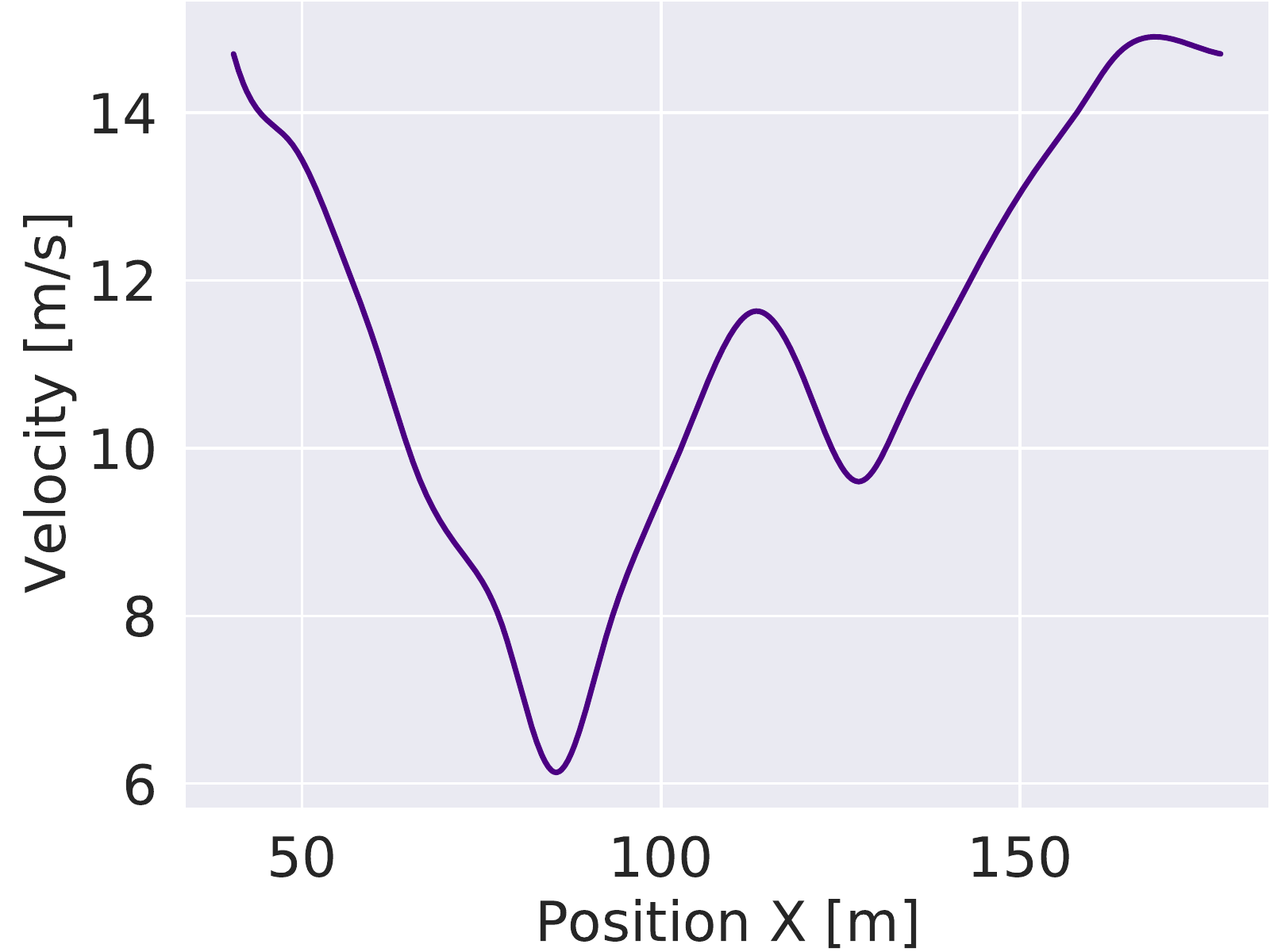}}
\subfloat[Heading angle]{\label{fig.heading2}\includegraphics[width=0.22\textwidth]{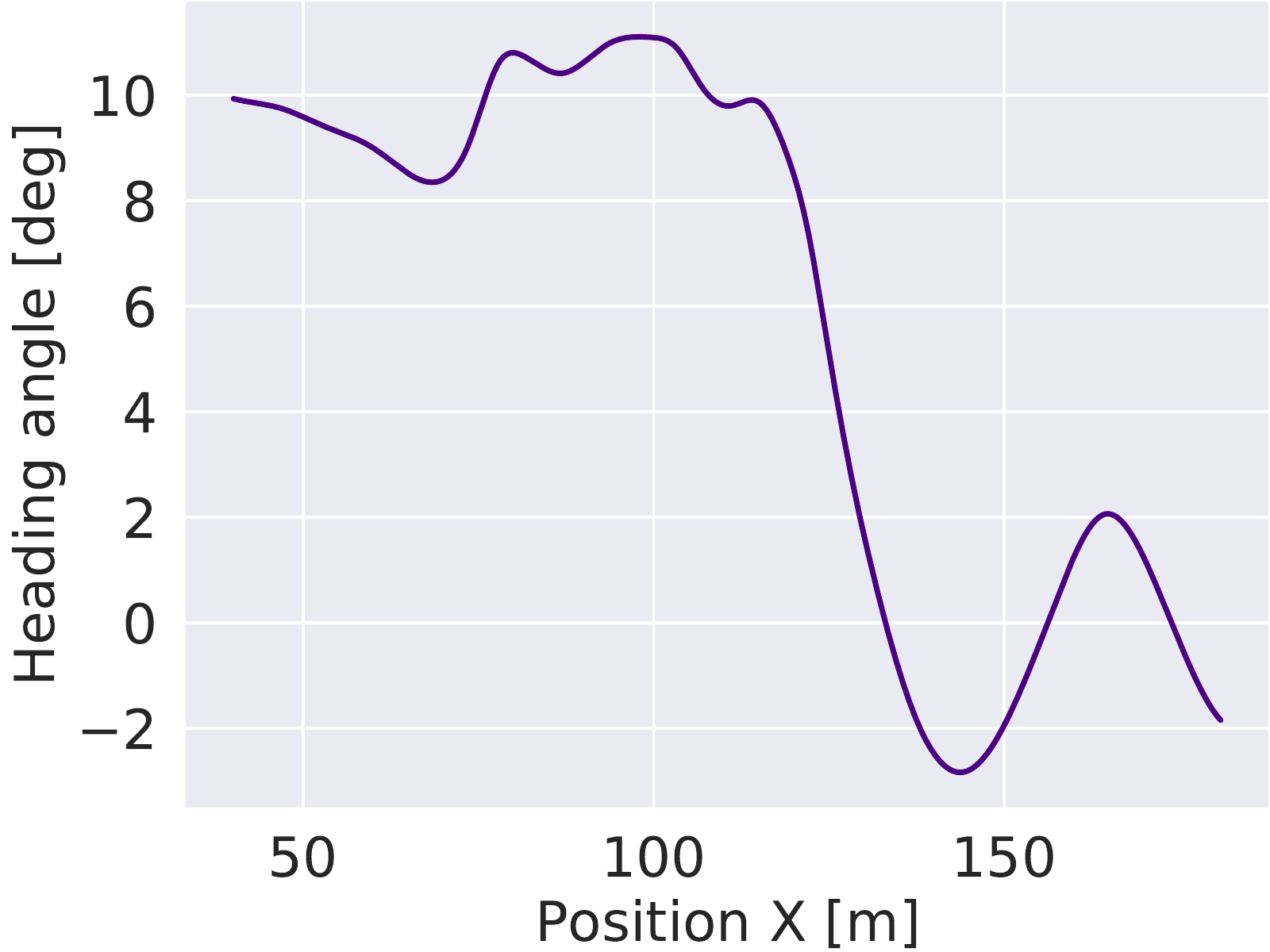}}\\
\subfloat[Acceleration]{\label{fig.acc2}\includegraphics[width=0.22\textwidth]{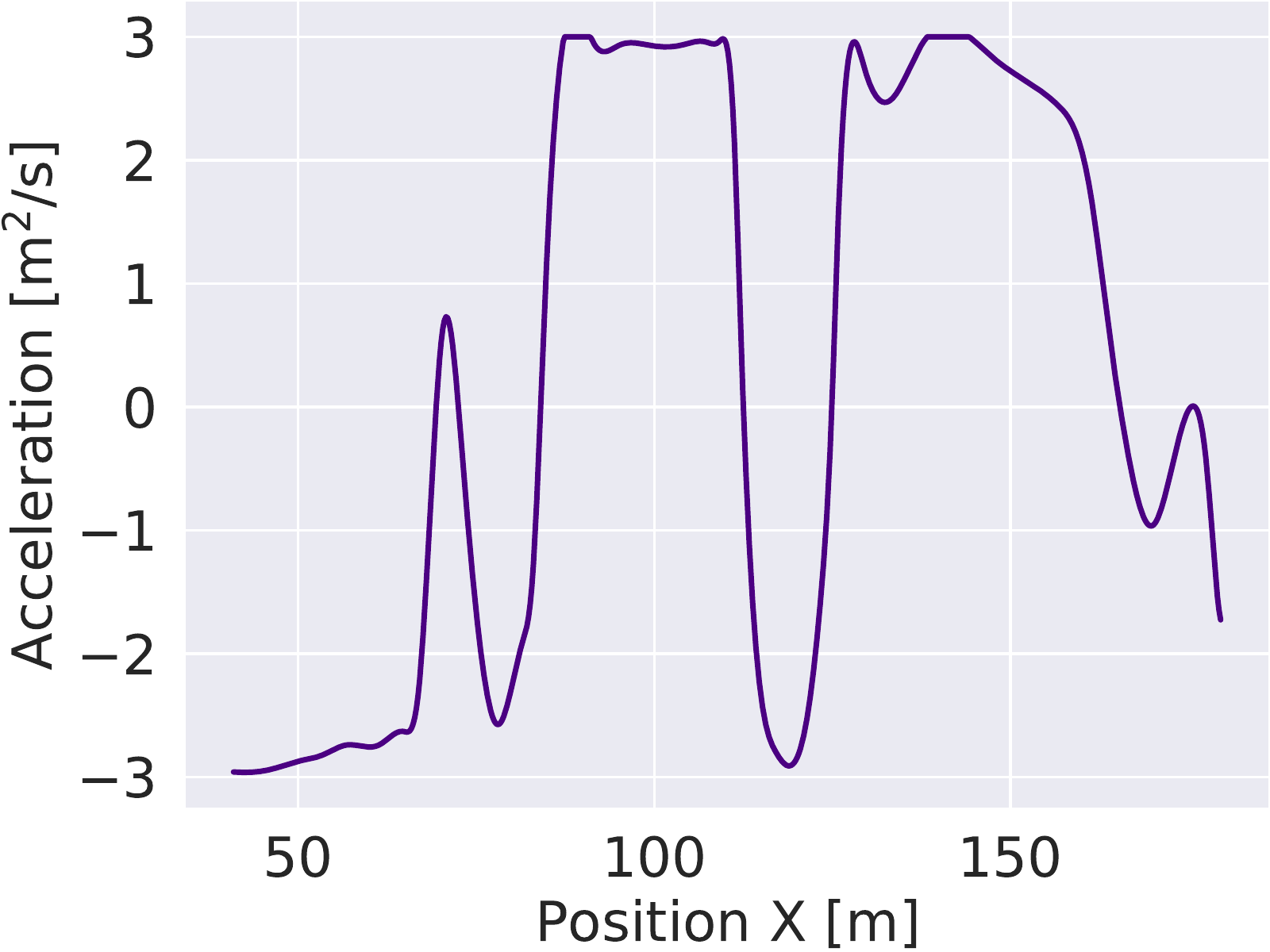}}
\subfloat[Front wheel angle]{\label{fig.delta2}\includegraphics[width=0.22\textwidth]{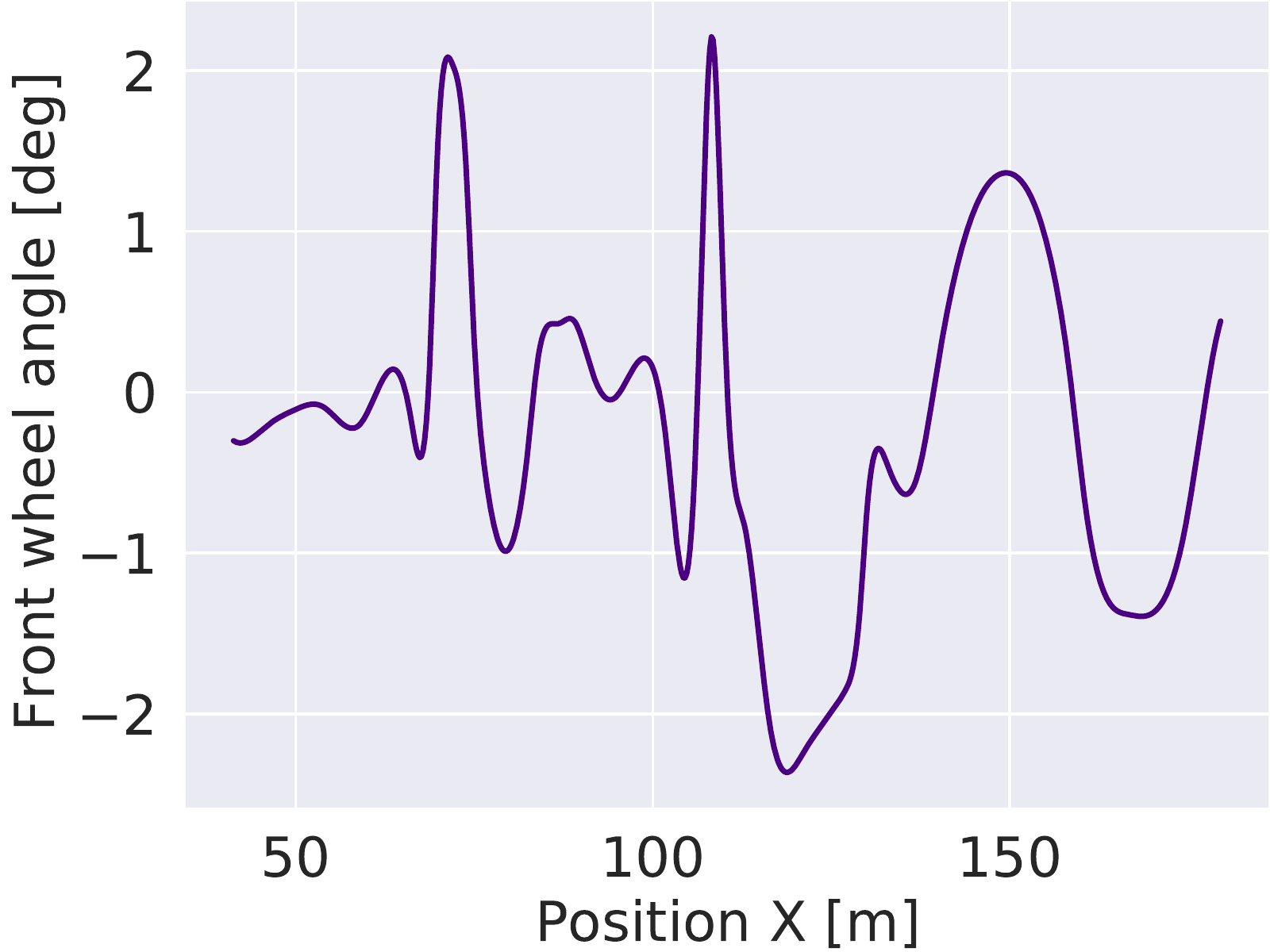}}
\caption{States and actions in simulation case 2.}
  \label{fig:Sim2_state_action}
\end{figure}

In the second simulation, the traffic density is set to be dense, as shown in Fig. \ref{fig:Sim2_Tra} and Fig. \ref{fig:Sim2_state_action}. At the start, the ego vehicle decelerates to keep away from the front vehicles for safety considerations. Once the ego enters the acceleration lane, it drives along the left side of the lane, heads to the mainline, and accelerates to find a gap for merging in. However, the mainline has no room for the merge because of traffic congestion. Therefore, the ego continues to drive in the acceleration lane. Then, two vehicles in the mainline turn right into the acceleration lane because of the traffic congestion, thus leaving a gap behind the ego. The ego then decelerate until the gap is large enough. Finally, the ego speeds up and merges in the mainline quickly before it hits the end of the acceleration lane.

These two cases show that the SDSAC exhibits intelligent driving behaviors to deal with the lone-term decision-making problems, generates diverse merging trajectories in different situations to make the automated vehicle drive safe and efficiently in the on-ramp merge scenario.

\section{Conclusions}
In this paper, we propose the Shielded Distributional Soft Actor-Critic (SDSAC) for safe and efficient decision-making under interactive on-ramp merge scenarios in an end-to-end way. The algorithm balances the performance of safety and efficiency by a framework of offline training and online correction, in which the policy evaluation with safety consideration and state constraints under barrier function condition are both adopted to support each other for better safety performance. In the offline training, the reward is designed with a safety term so that the policy update is guided by a comprehensive evaluation. That reduces the reliance on the safety shield and then the probability of its failure. In the online correction, a safe action is computed from the output of the trained policy by minimizing its distance from the safe action space. To avoid infeasible problems, we control the 
boundary of the safe space using the barrier function technique. 
The statistical results suggest that the SDSAC achieves efficient driving while having the best safety performance compared to baseline algorithms. In addition, the learned driving policy generates diverse merge trajectories in different simulation settings to handle long term decision-making problems, verifying the effectiveness of the method. 


\printendnotes


\bibliography{main.bbl}



\end{document}